%% file: 00_main_file.tex
\documentclass[prodmode]{acmsmall}

\usepackage[ruled]{algorithm2e}

\usepackage{amsmath}
\usepackage{amssymb}

\DeclareTextFontCommand{\mytexttt}{\ttfamily\hyphenchar\font=45\relax}

\usepackage[table]{xcolor}
\usepackage{hhline}

\usepackage{mathtools}
\usepackage{todonotes}
\usepackage{graphicx}
\usepackage{caption}
\usepackage[export]{adjustbox}
\usepackage{subcaption}
\usepackage{pgfplots}
\usepackage{cite}
\usepackage{array}

\usepackage{adjustbox}

\usepackage{color,soul}
\usepackage[normalem]{ulem}
 
\usepackage{multirow}

\usepackage[hidelinks]{hyperref}

\definecolor{wikiblue}{HTML}{0645ad}
\definecolor{verylightgrey}{RGB}{224,224,224}

\DeclareRobustCommand{\hlgr}[1]{{\sethlcolor{verylightgrey}\hl{#1}}}

\newcommand{\wikilink}[1]{{\textcolor{wikiblue}{#1}}}
\newcommand{\wikilinkhl}[1]{{\textcolor{wikiblue}{\hlgr{#1}}}}
\newcommand{\wikifootnote}[1]{\textcolor{wikiblue}{\textsuperscript{[#1]}}}
\newcommand{\lang}{P}

\makeatletter
\newcommand*{\getcountref}[1]{%
	\expandafter\@getcountref\csname r@#1\endcsname
}
\newcommand*{\@getcountref}[1]{%
	\ifx#1\relax
	0
	\else
	\expandafter\@car#1\@empty\@nil
	\fi
}
\makeatother

\SetAlFnt{\small}
\SetAlCapFnt{\small}
\SetAlCapNameFnt{\small}
\SetAlCapHSkip{0pt}
\IncMargin{-\parindent}

\newcolumntype{P}[1]{>{\raggedright\arraybackslash}p{#1}}

\makeatletter
\def\@copyrightspace{\relax}
\makeatother

\begin{document}
	
\markboth{S. Gottschalk et al.}{MultiWiki: Interlingual Text
Passage Alignment in Wikipedia}

\title{MultiWiki: Interlingual Text Passage Alignment in Wikipedia}
	\author{SIMON GOTTSCHALK
		\affil{L3S Research Center, Hannover, Germany}
		ELENA DEMIDOVA\footnote{Corresponding author: Elena Demidova,
			demidova@L3S.de}
		\affil{University of Southampton, UK and L3S Research Center, Hannover,
			Germany}}


\begin{abstract}
	In this article we address the problem of
text passage alignment across 
interlingual article pairs in Wikipedia.
We develop methods that enable the identification and 
interlinking of text passages written in different languages 
and containing overlapping information. 
Interlingual text passage alignment can enable Wikipedia editors 
and readers to better understand language-specific context 
of entities, provide valuable insights in  
cultural differences and build a basis for qualitative analysis of the articles.
An important challenge in this context is the trade-off between the 
granularity of the extracted text passages and the precision of the alignment.
Whereas short text passages can result in more precise alignment, 
longer text passages can facilitate a better overview of the differences 
in an article pair.  
To better understand these aspects from the user perspective, 
we conduct a user study 
at the example of the German, Russian and the English Wikipedia and collect  
a user-annotated benchmark.
Then we propose MultiWiki -- a method that adopts an integrated approach to the
text passage 
alignment using semantic similarity measures and
greedy algorithms and achieves precise results with respect to the user-defined
alignment.
MultiWiki demonstration is publicly available and currently supports four
language pairs.
\end{abstract}

%
%


%
%



\begin{bottomstuff}	
	This work was partially funded by the ERC
under ALEXANDRIA (ERC 339233), H2020-MSCA-ITN-2014
WDAqua (64279) and COST Action IC1302 (KEYSTONE).

\end{bottomstuff}

\maketitle

\input{01_introduction}

\input{02_problem}

\input{03_para_similarity_function}

\input{04_para_matching}

\input{05_sentences}

\input{06_userstudy}

\input{07_paragraphs_evaluation}

\input{08_related}

\input{09_conclusion}

\bibliographystyle{ACM-Reference-Format-Journals}
\bibliography{00_main_file}

\received{July 2016}{November 2016}{November 2016}
	
\end{document}

%% file: 01_introduction.tex
\section{Introduction}
\label{sec:introduction}

Articles containing information about entities of public interest
become increasingly available in different
languages on the Web, within community-created
knowledge bases, encyclopedias and on the online news. 
As these sources evolve independently in each
language, they often reflect community-specific points of view
\cite{Rogers2013} and can contain 
complementary and sometimes contradictory information. 
This diversity is particularly interesting 
in the context of events influencing
several communities (e.g. the Brexit, the
refugee crisis in Europe and the Snowden affair).
In order to provide an overview of the language and community-specific facets of the
entities, help users to identify overlapping and complementary information
and enable quality control in multilingual datasets, methods for effective  
identification and interlinking of related information across languages are
required.

One prominent example of a large community-created interlingual data source is 
Wikipedia -- an online encyclopedia available in 
more than 290 language editions, counting above 50 million users from all 
over the world
and containing more than 30 million 
articles.\footnote{\label{note1}\url{https://meta.wikimedia.org/wiki/List_of_Wikipedias}}
In Wikipedia, the articles representing equivalent real-world entities 
in different language editions become increasingly 
interlinked. 
In the following, we refer to such interlinked articles written in 
different languages as \textit{partner articles}. 
The independent evolution of the Wikipedia language editions often leads to 
significant semantic differences 
across the partner articles. 
For example, Table \ref{tab:introduction_example2} illustrates inconsistencies
in the German\footnote{\url{https://de.wikipedia.org/wiki/Gohi_Bi_Cyriac?oldid=136482800}} and English\footnote{\url{https://en.wikipedia.org/wiki/Gohi_Bi_Zoro_Cyriac?oldid=637509171}} versions of the article ``Gohi Bi Zoro Cyriac'' caused 
by the information contained only in the German version and related to the 
move of this footballer to Charlton Athletic in 2007. 
On a more general note, previous studies have shown the information asymmetries
across Wikipedia language pairs: Although the English Wikipedia is by far the biggest
with respect to the number of articles, edits and
users\footnotemark[\getcountref{note1}], it has been shown that for many entities,
Wikipedia articles in other languages are much longer than the corresponding 
descriptions in English and may contain
contradictory information \cite{Filatova:2009}. Paramita et al.
\citeyear{Paramita:2012} conducted a user study on a random sample of 800
cross-lingual partner articles to find out that 28.8\% of them are
only moderately similar and 18.8\% were judged to be different.

\begin{table}
\tbl{A user-aligned pair of text passages from the 
Wikipedia article ``Gohi Bi Zoro Cyriac'' in the English and the German 
Wikipedia, along with a manual English translation of the German text 
passage. The highlighted information about a move of the football player to 
Charlton Athletic in 2007 is only present in the German 
version of the article.\label{tab:introduction_example2}}{%
	\centering
	\begin{tabular}{P{4.3cm}|P{4.2cm}|P{4.2cm}}
	\textbf{English Text Passage} & \textbf{German Text Passage} & \textbf{German Text Passage (Translated)} \\ \hline
	\scriptsize
	\fontfamily{pcr}\selectfont
	In March 2004, he joined \wikilink{ASEC Mimosas}.\wikifootnote{2}\newline 
	Cyriac was a topscorer of \wikilink{C\^{o}te d'Ivoire Premier Division} 
	in 2008 season.\newline On 31 January 2009, he moved 
	to \wikilink{Standard Li\`{e}ge} signing a five-year contract with Belgian champions.
	& 	\scriptsize
	\fontfamily{pcr}\selectfont 2004 wechselte er in die zweite Mannschaft von \wikilink{ASEC Mimosas}, \hlgr{welche er 2007 Richtung }\wikilinkhl{England}\hlgr{ verlie{\ss}}. \newline \hlgr{Der Ivorer unterschrieb bei }\wikilinkhl{Charlton Athletic}\hlgr{, jedoch wurde er von den Addicks an seinen Jugendverein weiterverliehen.} Mit den ASEC wurde er Vizemeister, zudem wurde er Torsch\"{u}tzenk\"{o}nig der h\"{o}chsten ivorischen Liga. \newline Nach einem weiteren halben Jahr in der Heimat wurde er im Januar 2009 von \wikilink{Standard L\"{u}ttich} verpflichtet. & 	\scriptsize
	\fontfamily{pcr}\selectfont In 2004 he changed over to the second team of
	\wikilink{ASEC Mimosas}, \hlgr{which he left towards }\wikilinkhl{England}\hlgr{ in 2007.} \newline \hlgr{The Ivorian signed at }\wikilinkhl{Charlton Athletic}\hlgr{, however the Addicks immediately loaned him to his youth club.} With ASEC he became vice champion, in addition he became topscorer of the highest Ivorian league. \newline After a further half a year in his home country, he was signed by \wikilink{Standard Li\`{e}ge} in January 2009.
	\end{tabular}}
\end{table}

Precise alignment of text passages containing overlapping 
information in partner articles 
can enable users to obtain a comprehensive overview over common
entity facets shared across the language editions
and their language-specific context.
On the one hand, manual alignment of overlapping information 
like it was performed in \cite{Rogers2013}
can be very precise. However, such manual alignment requires user proficiency in 
a foreign language
and can be a very time consuming and daunting task even for an expert user,
especially for longer articles.
On the other hand, an automatic alignment of semantically overlapping 
text passages is a challenging problem. This is due to the 
varying granularity of the overlapping text passages, differences in the text
flow, additional information (facts or intermediate sentences) 
that does not have a direct correspondence in the other language 
as well as different linguistic structures used to express 
equivalent information.
Therefore, it is important to develop automatic methods that are able to 
identify semantically 
similar text passages, while being robust against syntactic 
differences.

We address the problem of the interlingual text passage 
alignment in order to facilitate a
comprehensive overview of the information shared by partner articles. 
Existing approaches to interlingual text alignment are limited to few
specific applications such as alignment of parallel fragments and sentences
to support machine translation (e.g. \cite{Fragment}, \cite{Mohammadi10})
and 
detection of plagiarism cases (e.g. \cite{Perez:2015}). 
On the one hand, fragment and sentence alignment fail to provide an overview of the 
overlapping article
parts due to their high granularity. On the other hand, 
interlingual plagiarism detection enforces strict conditions on the 
overlapping parts and is therefore not directly applicable to align text
passages providing complementary information and having significant structural
differences.
Overall, existing text alignment methods are not suitable to provide a
comprehensive overview of the interlingual article overlap.

In this article, we present a novel approach to 
\textit{interlingual text passage alignment} across 
partner articles. 
To facilitate this alignment, we rely on
semantic information including overlapping entities,
time expressions related to common time intervals and selective terms. 
Text passages are extracted concurrently
from partner articles
and aligned based on their interlingual semantic similarity, 
while simultaneously enforcing granularity-related objectives and taking into
account the interlingual context.
In summary, the contributions of this article are as follows: 
 (i)  We present the problem of interlingual 
  text passage alignment. To the best of our knowledge, this problem is not  
addressed by any existing approach;
  	(ii)  We propose an effective method for interlingual 
	text passage alignment based on semantic similarity measures and greedy algorithms;
	 (iii)  We conduct a user study and create a user-annotated benchmark. 
We make this
benchmark publicly available to encourage further research in this area.
Our experiments demonstrate that the proposed method 
is effective with respect to both, precision of the alignment and granularity of the extraction.
Text passages aligned by our method closely match the user-defined annotations.

MultiWiki demonstration is currently
available in four language pairs: German-English, Dutch-English, Portuguese-English and
Russian-English.\footnote{http://multiwiki.l3s.uni-hannover.de/demo.html}
In this article we provide evaluation results for German-English and
Russian-English pairs.
The MultiWiki text passage alignment method presented in this article 
can facilitate a wide range of interlingual applications. 
For example, in our recent demo paper \cite{gottschalk2016}
we presented a novel application to 
analyze the interlingual temporal evolution of the article pairs.
This application enables users to observe
the propagation of information across the language editions of the
articles 
on a timeline and to perform
a detailed visual comparison of the article snapshots 
at a particular point in time.
Using the 
MultiWiki text passage alignment method presented in this article we can
facilitate an effective visual
comparison of the partner articles in this application.

The rest of this article is organized as follows: In Section
\ref{sec:problem_and_overview} we provide a formal definition of the 
problem of the interlingual text passage
alignment and provide an overview of our approach. 
Then, we present our
methodology to address this problem, including: 1) A semantic similarity
function described in Section \ref{sec:alignment_function};
and 2) An interlingual text passage alignment procedure presented in Section
\ref{sec:alignment_procedure}.
Following that, we describe the fine-tuning of the similarity
function and its evaluation in Section
\ref{sec:sim_function_tune}.
In order to further fine-tune our interlingual text passage alignment method and
to enable its evaluation, we conduct a user
study presented in Section \ref{sec:userstudy} and collect a
user-annotated benchmark.
Following that, we discuss the evaluation results of the text passage alignment
method on German-English and Russian-English article pairs
in Section \ref{sec:evaluation}. Section \ref{sec:related} provides a
related work overview. Finally,
we discuss our contributions, limitations of the approach and future plans in
Section \ref{sec:discussion}.

%% file: 02_problem.tex
\section{Problem Statement and an Overview of the MultiWiki Approach}
\label{sec:problem_and_overview}

\begin{figure*}[!ht]
	\centering
	\includegraphics[width=1.0\textwidth]{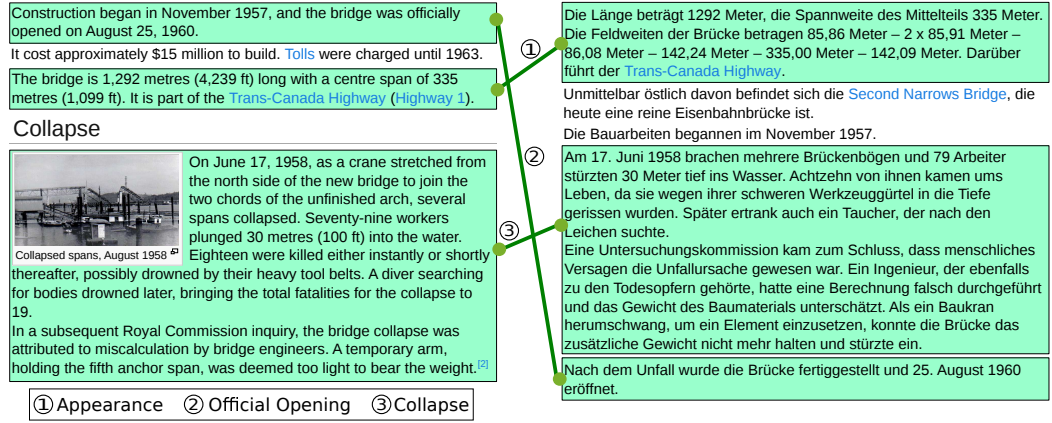}
	\caption[Example]{A user-annotated example illustrating an extract 
		from the partner articles
		entitled ``Ironworkers Memorial Second Narrows Crossing'' from the English
		and German Wikipedia language
		editions (as of the 1st October, 2015).
		Aligned text passages are enclosed by the bounding boxes, highlighted in green,
		connected via the green lines and manually annotated. Included photography: \copyright{ }Ron B. Thomson, licensed under CC BY-SA 3.0\footnotemark.}
	\label{fig:introduction_example}
\end{figure*}

The goal of the \textit{interlingual text passage alignment} is to
extract and align text passage pairs containing overlapping 
information from partner articles
to facilitate an effective overview of the interlingual similarities and 
differences across these articles.
To illustrate the interlingual text passage alignment from the user perspective,
in Fig. \ref{fig:introduction_example} we present a user-annotated example from the
partner articles ``Ironworkers Memorial Second Narrows Crossing'' -- representing
a bridge in Canada -- in the English\footnote{\url{https://en.wikipedia.org/wiki/Ironworkers_Memorial_Second_Narrows_Crossing?oldid=674828683}} and the German\footnote{\url{https://de.wikipedia.org/wiki/Ironworkers_Memorial_Second_Narrows_Crossing?oldid=130806835}} Wikipedia. This article pair 
has been manually annotated
by a user to identify overlapping interlingual text 
passages.\footnote{Our user study and benchmark creation is presented in Section
\ref{sec:userstudy} in more detail.}
In this example, the user manually identified and labeled three 
interlingual text passage pairs,
such as ``Appearance'', ``Official Opening'' and ``Collapse'' of the bridge in June 1958.

\footnotetext{\url{https://creativecommons.org/licenses/by-sa/3.0/deed.en}}
\subsection{Problem Statement}
\label{sec:problem}

In this section we first define the notions 
of a text passage, text passage alignment and the interlingual text passage
similarity.
Following that we discuss the interplay 
of the interlingual text passage similarity and the granularity of the
extraction.
Finally, we introduce the interlingual text passage alignment 
as an optimization problem.

A \textbf{Text Passage} is a 
non-empty list of 
consecutive sentences in an article.
In the context of text passage alignment, we assume that text passages are
topically coherent.

\begin{definition}[Text Passage]
Let $A_{\lang} =$ $(s_1,$ $\cdots,$ $s_N)$ represent an article $A_{\lang}$ from the
language edition $\lang$ through its sentence list $(s_1,\cdots,s_N)$, $N
\geqslant 1$.
Then, a \textit{text passage} $T_{\lang_j} \subseteq A_\lang$ is a consecutive non-empty
fragment of the sentence list $A_\lang$,
such that all sentences in $T_{\lang_j}$ are related to a common latent topic.
\label{def:TextPassage}
\end{definition}

\textbf{Text Passage Extraction and Alignment:} 
Text passage extraction (i.e. the identification of the text passage borders
within an article) and alignment (i.e.
the identification of the most relevant text passage in the partner
article) both depend on the interlingual context given by the partner
article.
To enable an efficient overview of the overlapping article parts, we define 
text passages in an article as mutually exclusive, i.e. containing
non-overlapping sentence sequences. 
The alignment of text passages is mutually
exclusive as well, i.e.
a text passage is aligned to the most 
relevant text passage in the
partner article.

\begin{definition}[Text Passage Alignment]
We use the notation $T_{\lang_j}\leftrightarrow T_{G_k}$ to represent the alignment of 
the text passages 
$T_{\lang_j} \subseteq A_\lang$ and $T_{G_k} \subseteq A_G$ from the partner articles 
$A_\lang$ and $A_G$, respectively.
The following conditions apply:
(1) Text passages in an article are mutually exclusive, i.e. a sentence
can belong to at most one text passage:
$~\forall T_{\lang_j}, T_{\lang_k} \subseteq A_{\lang}: T_{\lang_j} \bigcap T_{\lang_k} = \emptyset $;
(2) Text passage alignment is mutually exclusive, i.e. one text passage 
is aligned to at most one text passage in the partner article;
and
(3) All sentences in $T_{\lang_j}\leftrightarrow T_{G_k}$ are related to a common
latent topic.
\label{def:TextPassageAlignment}
\end{definition}

\textbf{Interlingual Text Passage Similarity:} 
An interlingual text passage pair is \textit{semantically similar} if these
text passages
share similar information nuggets. An
information nugget can be an entity, a fact, a similar piece of information, 
or an answer to a question \cite{Clarke:2008}.
The interlingual text passage similarity can be estimated using a similarity
function based on semantic and syntactic features.
The value of the similarity function should correlate 
with the overall similarity of the information nuggets contained in the text
passage pair.

\begin{definition}[Interlingual Similarity Function]
Let $T_{\lang_j} \subseteq A_\lang$, $T_{G_k} \subseteq A_G$ be two text passages in the 
partner articles $ A_\lang$ and $ A_G$, respectively.
Then, $Sim_{F}(T_{\lang_j}, T_{G_k})  \in [0,1] $ is the function that estimates
an interlingual similarity of these text passages
using a set $F=\{f_1, \ldots, f_N\}$ of semantic and syntactic features. 
$Sim_{F}$ is monotonically 
increasing, with ``1'' corresponding 
to the highest similarity. 
\label{def:ISimilarityFunction}
\end{definition}

\textbf{A Trade-off between the Similarity and Granularity of Aligned Text
Passages:}
One way to maximize the similarity of the aligned text passages is to increase 
their granularity (i.e. to extract text passages that contain less sentences). 
In an extreme case this naive approach results in a large number of short 
text passages (e.g. text passages consisting of a single sentence each). 
However, such high-resolution alignment fails to provide a comprehensive 
overview of the common facets covered in the article pair. 
At the other extreme, in case of a low-granularity alignment, an entire article
could be considered as one long text passage. Such alignment can likely result
in low similarity due to the potentially high proportion of 
dissimilar information nuggets
in the text passage pair, failing to meet the overview goal either.
Therefore, an effective text passage alignment method should concurrently
enforce the objectives related to the semantic similarity and the
granularity of extracted text passages.
\newpage
\textbf{The Objectives of the Alignment:}  
The \textit{interlingual text passage alignment} aims at the following
objectives:

\begin{itemize}
\item [1:] Maximize the similarity of the aligned text passages 
in an article pair. 
\item [2:] Minimize the overall number of the extracted text passages.
\end{itemize}

\subsection{An Overview of the Text Passage Extraction and Alignment in
MultiWiki}
\label{sec:overview}

Given the optimization problem 
of the interlingual text passage alignment defined in Section \ref{sec:problem},
our method relies on the two key components: 1) A semantic similarity
function that enables precise assessment of the interlingual text passage
similarity for text passages containing overlapping information
nuggets; and 2) A greedy algorithm that incrementally extracts similar text
passages from the interlingual article pairs using their mutual context 
to create an effective alignment.

To enable fine-tuning and evaluation of the proposed method, we create two
user-annotated benchmarks: 1) $Sim-B$ that provides 
continuous similarity values for text passage pairs at the sentence level and
thus facilitates efficient fine-tuning and evaluation of the
similarity function; and 2) $Align-B$ that
contains aligned interlingual 
text passage pairs extracted and annotated by the users to 
facilitate fine-tuning and evaluation of the text passage extraction and
alignment algorithm.
To facilitate further research in this area, our benchmarks are publicly 
available.\footnote{\url{http://multiwiki.l3s.uni-hannover.de/benchmark.html}}

%% file: 03_para_similarity_function.tex
\section{Interlingual Text Passage Similarity}
\label{sec:alignment_function}

In order to facilitate text passage alignment we need to estimate
an interlingual similarity of text passages by instantiating the similarity
function introduced in Definition \ref{def:ISimilarityFunction}. 
This function uses a set $F=\{f_1, \ldots, f_N\}$ of
semantic and syntactic features.
The choice of the features in this article is driven by two factors: 
1) The availability of interlingual translation services 
and extractors that enable effective and efficient 
extraction of feature values in the
interlingual settings; and 2) The intuition that the
features correlate with the overall interlingual text
passage similarity.

Intuitively, co-occurring selective terms and semantic annotations such as named
entities and time expressions can substantially contribute towards precise text
passage alignment, in particular in the case of partial information overlap. 
In order to facilitate computation of the term-based similarity, in this 
article
we use English as a pivot language due to the relatively high availability of
the translation services. 
In particular, in our experimental evaluation we use the Bing translation
API that enables high quality machine translation
in more than 50 languages.\footnote{https://www.microsoft.com/en-us/translator/translatorapi.aspx}
Semantic features such as named entities and time expressions can be efficiently
extracted and co-referenced in a number of languages using state-of-the-art tools such as 
DBpedia Spotlight \cite{Daiber2013} and HeidelTime \cite{StroetgenGertz2013}.
The feature set in our model is easily extendable, such that in
case further interlingual semantic information extractors become proficient,
new features can be added. For example, open relation extraction (e.g. in
\cite{FaruquiK15}) is an interesting direction to add more semantic 
information to the model in the future.

We assume a linear dependency between the features under consideration
and the overall text passage similarity. 
The motivation for the linear combination is its simplicity, efficiency of 
training and computation as well as its effectiveness, as demonstrated by our results.
Therefore, we model the
similarity function as a linear combination. Using this
modeling, feature weights can be
efficiently learned from annotated datasets, e.g. using 
linear regression.
The importance of the features in the context of the interlingual text passage
alignment is represented using the \emph{feature importance factors} (or weights) $\beta_i \in [0,1]$,
with $\sum_i \beta_i = 1$:

\begin{equation}
\label{eq:al_function}
\begin{split}
Sim_{F}(T_{\lang_j}, T_{G_k}) =
\sum_{i=1}^N \beta_i \times {sim}(T_{\lang_j}, T_{G_k}, f_i),
\end{split}
\end{equation}


\noindent where ${sim}(T_{\lang_j}, T_{G_k}, f_i)\in [0,1]$ is the similarity of text 
passages $T_{\lang_j}$ and $T_{G_k}$
computed using feature $f_i$. 

\subsection{Text Passage Alignment Features}
\label{sec:features}

In the following, we present the features that we found to be effective 
for the interlingual text passage alignment 
and the corresponding similarity computation in more detail. 

\textbf{Cosine Similarity (Co)}: 
\textit{Cosine Similarity} measures the similarity of text passages using 
terms translated to a pivot language, 
while taking term frequency ($tf$) and selectivity ($idf$) of the terms into
account \cite{manning2008}.
To increase the precision of the alignment, 
the terms are pre-processed using stemming and stop word removal. 
Finally, the text passages are represented as vectors of $tf-idf$ term weights
and the cosine similarity of the vectors is computed. 
Using \textit{Cosine Similarity} text passage pairs containing selective terms, 
i.e. the terms that
can distinguish a particular text passage pair from the rest of the corpus, are
prioritized.

Text passage alignment using \textit{Cosine Similarity} taken in isolation
works well for parallel text passages that contain equivalent information. 
For example, this can be the case if one article is created as a translation
of the other. In case of the partial overlap, term-based alignment is not
sufficient to distinguish between semantic similarity, i.e. common information
nuggets, and simple overlap in selective terms.
Therefore, we do not expect \textit{Cosine Similarity} taken in isolation to
precisely distinguish between text passage pairs containing common information
nuggets and text passage pairs containing parallel fragments. 
In order to enable for precise interlingual alignment of partially overlapping
text passages, we use further features such as \textit{Entity Annotations} and
\textit{Time Annotations}.

\textbf{Entity Annotations (E)}: \textit{Entity Annotations} are references 
to named entities mentioned in the text passages.
Named entities are one of the key semantic components to support an
effective alignment of text passages containing common 
information nuggets across languages. 
In order to enable effective usage of 
\textit{Entity Annotations} for text passage alignment,  
interlingual entity co-referencing 
and sparsity of entity annotations need to be addressed.

Within a particular language named entity references can be extracted and
co-referenced using existing annotation tools 
(e.g. DBpedia Spotlight \cite{Daiber2013}).
Also, interlingual annotation tools such as Babelfy become recently available
\cite{Moro:2014}.
In this work, 
we annotate the entities using DBpedia
Spotlight in the original language versions of the articles and then establish 
interlingual links between the \textit{Entity Annotations}
using Wikipedia language links (i.e. the links connecting partner articles in
Wikipedia).
Another possible solution would be to use a machine translation service before
named entity disambiguation is applied. However, we observed that machine translation services
(such as Bing Translator API) often fail to correctly translate named entity
labels.

Due to the sparsity and the distribution of 
the \textit{Entity Annotations} in text passages,
directly applying cosine similarity measure
to these annotations does not
lead to a very precise text passage alignment.
For example, if two text passages in the English and the German article 
``Japan'' only have a single \textit{Entity
	Annotation} ``Tokyo'' each, their similarity shall not be very high 
	because of the high frequency of this
annotation within the article. However, cosine similarity
returns the maximum similarity of $1$ because the vector representations
of the text passages are identical in this case. 
This problem has also been observed
in the related approaches that use sparse annotations and short texts
as features (e.g. in \cite{Duh:2013}).
Therefore, our approach is to put an additional emphasis on the 
highly selective entity annotations.
To this extent, we compute the text passage similarity using the cosine of the
vectors containing annotations and
add a smoothing factor $\vec{n}$, further emphasizing 
selectivity of the entities in an article pair.

For each \textit{Entity Annotation}, the smoothing factor
shall be equal to $1$ if this annotation is unique in both articles and 
shall approximate $0$ if the annotation is very frequent. 
Moreover, the function should quickly 
decrease with the decreasing selectivity of the annotations.
These conditions are fulfilled by a function for exponential decay. 
Therefore, we create a vector $\vec{n}$, where $n_i$ is the weight of 
the \textit{Entity Annotation} $i$ computed as:
\begin{equation}
\label{eq:smoothing}
n_i = e^{- \frac{df_{i}}{\alpha}},
\end{equation}
\noindent where $df_i$ denotes the number of sentence pairs (i.e. the shortest
text passages) in the article pair containing the annotation $i$.
Note that this smoothing factor does not take the length 
of the document into account.
The weights computed by Equation \ref{eq:smoothing} are in the interval 
$n_i \in ($0$, $1$]$ with the lower weights corresponding to the more 
common annotations. 
The greater $\alpha$, the slower the decay: If $\alpha$ is very large, 
highly frequent annotations are 
assigned a greater smoothing value and vice versa.
With $\alpha=12.2$ we maximized the correlation between the 
proposed similarity function and the similarity derived from the user
annotations on the training dataset $TD_1$ described later in Section
\ref{sec:sim_function_tune} as measured using the Pearson 
Correlation Coefficient (PCC).

When calculating the similarity $sim_E(T_{\lang_j},T_{G_k})$ of two text 
passages $T_{\lang_j}$ and $T_{G_k}$ based on 
the \textit{Entity Annotations}, $tf-idf$ weights
of the annotations are adjusted by $\vec{n}$:
\begin{equation}
\label{eq:entitysim}
sim_E(T_{\lang_j},T_{G_k}) = \frac{\sum_{i=1}^N w_{i,T_{\lang_j}} w_{i,T_{G_k}}n_i}{\sqrt{\sum_{i=1}^N w_{i,T_{\lang_j}}^2}\sqrt{\sum_{i=1}^N w_{i,T_{G_k}}^2}},
\end{equation}
\noindent where $w_{i,s_j}$ is the $tf-idf$ weight of the annotation $i$ in the
text passage $s_j$ and $N$ is the number of distinct aligned annotations in both
articles.

\textbf{Time Annotations (T)}: 
\textit{Time Annotations} are normalized time expressions extracted 
from text passages. These annotations 
are an important factor to support an
effective interlingual alignment, in
particular with regard to the temporal facts.
As Wikipedia is an encyclopedic text collection, time expressions
play an important role: In a subset of English and German
Wikipedia articles, we observed that on average 
$36$\% of the sentences contain time expressions.
When analyzing time expressions mentioned in the text passages
it is important to take their semantic similarity into account.
First, extraction and normalization of time expressions allows a more accurate 
comparison of the described time intervals than a pure syntactic similarity.
Second, time expressions describing longer time intervals, such as a year or a  
month are less precise and thus contribute less to the overall text passage similarity 
than more concrete time points such as a date. 
Thus, it is important to enable an accurate comparison of the (partially) overlapping time intervals 
taking the length of the interval into account.

Therefore, we assign relevance values to the time intervals $ta$ according 
to their length:
The longer the time interval, the smaller the relevance value.
Following this intuition, we set the weight of the time 
interval $w(t_i) = 1$, if $t_i$ represents a particular date, 
$w(t_i) = 0.85$ for a month and $w(t_i) = 0.6$ for a year.
We experimentally observed that the function configuration using these weights
outperforms other configurations, such as equal weights 
for the time
intervals of different length.

To compute the similarity based on the \textit{Time Annotations} $sim_T(\allowbreak T_{\lang_j},\allowbreak T_{G_k})$ 
between two text passages $T_{\lang_j}$ and $T_{G_k}$, 
we align each \textit{Time Annotation} $t_{i} \in ta_{T_{\lang_j}}$ with its best matching counterpart $t_{j} \in ta_{T_{G_k}}$ (if any) 
in these text passages and sum up the minimum relevance values of the aligned annotations to obtain
a time overlap value $tovl$:
\begin{small}
	\begin{equation}
	\begin{split}
	\label{eq:timesim:ovl}
	tovl (T_{\lang_j},T_{G_k}) = \sum_{t_{i} \in ta_{T_{\lang_j}}} \sum_{t_{j} \in ta_{T_{G_k}}}
	\begin{cases}
	\text{min} (w(t_i), w(t_j))~^*\\
	0,~otherwise
	\end{cases} 
	\end{split}
	\end{equation}
\end{small}
\noindent$^*$if $t_{i}$, $t_{j}$ refer to an overlapping time interval,
and there is no other overlapping $t_{j'} \in ta(T_{G_k})$ with a higher 
weight for $\text{min} (w(t_i), w(t_{j'}))$.

If, for example, the annotations ``2011/\allowbreak03/20'' and 
``2011/03'' are aligned, the relevance weight for a month is taken.
The time overlap is computed for both directions, summed up and then normalized by the total number 
of the \textit{Time Annotations} $|ta_{T_{\lang_j}}|+|ta_{T_{G_k}}|$ in the text passages $T_{\lang_j}$ and $T_{G_k}$:
\begin{small}
	\begin{equation}
	\begin{split}
	\label{eq:timesim}
	sim_T(T_{\lang_j},T_{G_k}) = 
	\frac{tovl(T_{\lang_j},T_{G_k}) + tovl(T_{G_k},T_{\lang_j})}{|ta_{T_{\lang_j}}|+|ta_{T_{G_k}}|}.
	\end{split}
	\end{equation}
\end{small}
For example, if a text passage $T_{\lang_j}$ contains 
the \textit{Time Annotations} ``2011/\allowbreak03/20'' and ``2011'' and 
text passage $T_{G_k}$ contains ``2011/03'', the similarity is calculated 
as $sim_T (\allowbreak T_{\lang_j},\allowbreak T_{G_k})=\frac{(0.85+0.6)+(0.85)}{2+1}\approx0.767$.

\subsection{Feature Extraction Pipeline}
\label{sec:pipeline}

\begin{figure*}[t]
	\centering
	\includegraphics[width=0.9\textwidth]{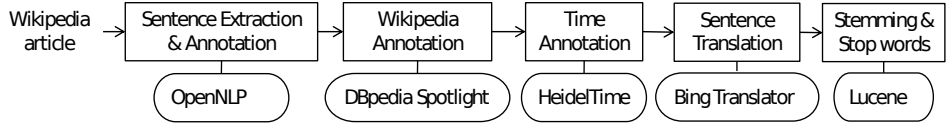}
	\caption{Processing Pipeline for Feature Extraction}
	\label{fig:pipeline}
\end{figure*}

In order to facilitate interlingual text passage alignment 
we apply a processing pipeline including the following steps: Sentence
splitting, entity annotation and annotation of time
expressions in the original language; Sentence translation to a pivot language; 
Stemming and stop word removal from the translated
sentences.
This easily reproducible pipeline is implemented using state-of-the-art 
tools and services and is presented in Fig.~\ref{fig:pipeline}.

\subsection{Feature Weights}
\label{sec:weights}

In order to obtain the feature weights, we created a benchmark $Sim-B$
described in Section \ref{sec:sim_function_tune} and performed function tuning.
As a result, our similarity function (\textit{CoET}) has the following feature
weights:
$\beta_{Co}=0.69$, $\beta_T=0.2$ and $\beta_E=0.11$.
As to not penalize the sentences with missing features, we have trained
additional similarity function configurations for these cases: \textit{CoE}
($\beta_{Co}=0.89$, $\beta_T=0$ and $\beta_E=0.11$), \textit{CoT}
($\beta_{Co}=0.8$, $\beta_T=0.2$ and $\beta_E=0$) and \textit{Co}
($\beta_{Co}=1.0$, $\beta_T=0$ and $\beta_E=0$).

%% file: 04_para_matching.tex
\section{The Alignment Procedure}
\label{sec:alignment_procedure}

The alignment of text passages is an optimization problem
that strives for high similarity of the extracted text passages and their 
low granularity simultaneously. 
The brute-force approach to this problem is not feasible 
as the number of possible text passages and their alignments 
grows exponentially with the number of sentences in an article pair.
Therefore, we propose a greedy approximation algorithm.

Intuitively, our method works bottom-up as follows: 
The algorithm starts with the alignment of the seed sentences in the partner
articles with the similarity above a pre-defined threshold $th$. Then, it
iteratively expands the alignment.
As long as the similarity $Sim_{F} (T_{\lang_j}, T_{G_k})$ 
of a text passage pair can be
increased by extending it with a close-by text passage pair, we
merge them, such that the overall similarity 
of the aligned text passages increases 
and the granularity of the aligned text passages decreases. 
The overall similarity is measured as 
$\sum_{T_{\lang_j}\leftrightarrow T_{G_k}\in S_{A_\lang,A_G}}{Sim_{F}
(T_{\lang_j}, T_{G_k})}$,
where $S_{A_\lang,A_G}$ is the set of all aligned text passages in
the article pair $A_{\lang}, A_{G}$.

Due to the interlingual differences, aligned text passages can contain different number of sentences.
 Consequently, it does not suffice to merge directly neighbored text passage pairs. Therefore, we 
propose two options to incrementally extend a text passage pair:

\begin{itemize}
	\item Merging with neighbored sentences: One of the text passages in a pair can be extended by a single neighbored sentence.
	\item Merging with (nearly) neighbored text passage pairs: Two text passage pairs can be merged if they are located in a close neighborhood. This implies the inclusion of intermediate sentences if the text passage pairs are not directly adjacent.
\end{itemize} 

\noindent Based on an initial alignment of similar sentences, we propose a greedy algorithm that extends the currently most similar text passage pair 
in each step until no extension is possible any more (i.e. 
until no text passage pair can be merged with 

To extract topically coherent text passages we rely on two estimates:
1) Interlingual context: Due to the different structure of the partner articles, text passages related to different topics are unlikely 
to come in the same order; Therefore if by adding more sentences to a text passage the interlingual similarity drops, 
the topics are likely to drift.
2) Text structure: Wikipedia articles are arranged in a
hierarchical structure. Articles consist of \textit{sections}
that may contain sub-sections, sub-subsections and
so forth. At the lowest level of the hierarchy, the text is split into
\textit{paragraphs}. The end of such (sub-)sections or paragraphs provides an 
indication of a possible topic drift; We introduce a parameter (Structure freedom ($sf$)) that 
allows different degrees of freedom with respect to this structure. 

In the rest of this section, we explain the merging methods in more detail 
and then show how the algorithm utilizes 
these methods to create an alignment of text passages.

\subsection{Merging of Text Passage Pairs with Neighbored Sentences}

The information nuggets of one sentence can be scattered over a few sentences in
the partner article, such that it is necessary to merge the sentences 
from the partner article to perform the alignment. 
Fig. \ref{fig:introduction_example} contains an example where this merging step is 
necessary: While the English article 
describes the bridge's collapse and the plunge of the workers in two sentences, 
the German article sums that information 
up just in one sentence: ``Am 17. Juni 1958 brachen mehrere Br{\"u}ckenb{\"o}gen
und 79 Arbeiter st{\"u}rzten 30 Meter tief ins Wasser.'' (Translated: ``On June 17, 1958, several bridge arches collapsed and 79 workers plunged 30 meters into the water''). In this case, 
the content of the German sentence is scattered among two directly adjacent English sentences. Thus, the English sentences are merged to 
form a larger text passage that can be aligned to the single German sentence.
More formally, if two consecutive sentences $s_i \in A_\lang$ and $s_{i+1} \in A_\lang$ both (partially) overlap with 
the same sentence $s_j \in A_G$ in the other article, 
they can be merged into the text passage $(s_i, s_{i+1}) \subseteq A_\lang$,  
which is aligned to form the text passage pair $(s_i, s_{i+1}) \leftrightarrow (s_j)$.

In Algorithm \ref{alg:para_matching}, the procedure
\mytexttt{mergeWithSentences} takes a text passage pair as an input and searches 
through all neighbored and unaligned sentences to merge them with it. When multiple text passages 
can be created that way, the one with the highest similarity $Sim_F$ is returned. 

\subsection{Merging Nearly Neighbored Text Passages}

Until now, we discussed the alignment with the help of single sentences. This is especially important 
at the beginning of the procedure to obtain a starting point for text passage extraction: 
Such aligned sentences constitute initial text passage pairs.
In the next step, we merge text passage pairs in a close neighborhood.
Due to the goal of similarity maximization, this merge may only be done if the similarity of the resulting 
text passage pair exceeds the similarity of the initial text passage pair. 
This condition ensures 
that the alignment simultaneously strives for low granularity and high overall similarity.
Under that condition, two text passage pairs $T_{\lang_{i}}  \leftrightarrow T_{G_{k}}$ and $T_{\lang_{j}} \leftrightarrow T_{G_{l}}$
can be merged, such that they are replaced by a single text passage 
pair $T_{\lang_{i'}} \leftrightarrow T_{G_{k'}}$, where $T_{\lang_{i'}}$ contains the sentences 
from $T_{\lang_{i}}$, $T_{\lang_{j}}$ 
and, potentially, the \textit{intermediate sentences} between them.
Such intermediate sentences can provide complementary information and 
while included within the aligned text passages, 
they are put in context.

Whether the similarity score can grow by merging text passages depends on how similar 
information is fragmented in both languages. In cases where there is no 1:1 correspondence 
at the sentence level, merged text passages help to better assimilate fragmented parts to match 
the information available on both sides, overall resulting in higher similarity after merging. 
Although the inclusion of intermediate sentences in text passages can potentially result in 
lower similarity values, it is not necessarily always the case. As long as the similar parts 
in the merged text passages overweight, similarity values of the resulting alignment will be higher.

We enable this extension method by the function \mytexttt{mergeWithPassagePair}: Given
a text passage pair $T_{\lang_{i}}  \leftrightarrow T_{G_{k}}$, all text passage pairs 
$T_{\lang_{j}} \leftrightarrow T_{G_{l}}$ in the close neighborhood 
 are chosen as
candidates and the one that results in the highest similarity after merging
is returned.

\subsection{The Alignment Algorithm}
\label{sec:algorthm}

With the help of the two merging functions \mytexttt{mergeWithSentence} 
and \mytexttt{mergeWithPassagePair}, 
we now define our algorithm \textit{MultiWiki} to extract and align a precise
and low-granular set of text passage pairs in a bottom-up manner. 
As shown in Algorithm \ref{alg:para_matching}, the input is 
are two articles, $A_{\lang}, A_{G}$, the similarity threshold $th$,
and the structure freedom parameter $sf$.

\begin{algorithm}[t]
	\SetAlgoNoLine
	\LinesNumbered

	\KwIn{Articles $A_{\lang},~A_{G}$, similarity threshold $th$, structure freedom
	parameter $sf$.}
	\KwOut{The set $S_{A_{\lang},~A_{G}}$ of the aligned text passages.}
	$S_{A_{\lang}, A_{G}}$ = $alignSentences(A_{\lang}, A_{G}, th)$\;
	$foundChanges \coloneqq$ \textbf{true}\;
	\While{$foundChanges$} {
		$foundChanges \coloneqq$ \textbf{false}\;
		sort($S_{A_{\lang},~A_{G}}$)\label{alg:build_paragraphs:sort}\;
		\For{each $T_{\lang_{i}} \leftrightarrow T_{G_{j}}$ in $S_{A_{\lang},~A_{G}}$ 	\label{alg:build_paragraphs:iterate}} {
			\If{\textbf{not} $mergedWithSentence(T_{\lang_{i}} \leftrightarrow
			T_{G_{j}})$ \label{alg:merged}} { $T_{\lang_{i'}} \leftrightarrow
			T_{G_{j'}} =$ mergeWithSentence($T_{\lang_{i}} \leftrightarrow
			T_{G_{j}}$,~$sf$)\label{alg:build_paragraphs_aggregate}\; } \Else { $T_{\lang_{i'}} \leftrightarrow T_{G_{j'}} =$ mergeWithPassagePair($T_{\lang_{i}}
				\leftrightarrow T_{G_{j}}$,~$sf$)\label{alg:build_paragraphs_proximate}\; }
				} \If{$Sim_F(T_{\lang_{i'}}, T_{G_{j'}}) > Sim_F(T_{\lang_{i}}, T_{G_{i}})$\label{alg:similarity_condition}} {
		$foundChanges \coloneqq$ \textbf{true}\;
		$S_{A_{\lang}, A_{G}} = (S_{A_{\lang}, A_{G}} \cup T_{\lang_{i'}} \leftrightarrow T_{G_{j'}}) \setminus T_{\lang_{i}} \leftrightarrow T_{G_{j}}$\label{alg:build_paragraphs_update}\;
		\textbf{break}\;
	}
}
\textbf{return} $S_{A_{\lang}, A_{G}}$\;
\caption{Text Passage Alignment Algorithm}
\label{alg:para_matching}
\end{algorithm}

In order to speed up the alignment process, the algorithm operates in a greedy manner: 
In each step it selects the currently most similar text passage pair $T_{\lang_{i}} \leftrightarrow T_{G_{j}}$ 
(line \ref{alg:build_paragraphs:sort} - line \ref{alg:build_paragraphs:iterate}).
$T_{\lang_{i}} \leftrightarrow T_{G_{j}}$ is either 
merged with a neighbored sentence or -- if this was already tried --
with a neighbored text passage pair (lines \ref{alg:merged} - \ref{alg:build_paragraphs_proximate}). 
If the extension is successful and the merged text passage pair $T_{\lang_{i'}} \leftrightarrow T_{G_{j'}}$
achieves a higher similarity score than $T_{\lang_{i}} \leftrightarrow T_{G_{j}}$, 
the set of text passage pairs is updated accordingly (line \ref{alg:similarity_condition} - 
\ref{alg:build_paragraphs_update}). This procedure terminates when no text passage pair can be extended
such that its similarity increases
: In this case, $S_{A_{\lang}, A_{G}}$ remains unchanged and 
the parameter $foundChanges$ stays \mytexttt{false}.

\subsection{Parameters}
\label{sec:parameters}

There are two parameters that can be tuned to adjust the behavior of the proposed algorithm
to make it better fit user preferences:

\begin{itemize}
    	\item \textbf{Similarity threshold ($th$):} A threshold 
	value $th$ that determines if a text passage pair is 
	regarded as being similar. A lower threshold enables more flexibility in the merging, 
	but can also affect the precision of the alignment.
	\item \textbf{Structure freedom ($sf$):} The likelihood of a 
	topic drift shall 
	be higher when reaching a new section or paragraph
	in the original text.
	Thus, to enhance the topical coherence of the aligned text passages, we can
	disallow the algorithm to merge sentences from different sections or 
	paragraphs. We introduce three structure 
	freedom levels: \textit{max} (i.e. no limits), \textit{mid} 
	(i.e. never exceed a given section)
	and \textit{min} (i.e. always stay within one paragraph). 
\end{itemize}

We discuss parameter tuning and their influence on the overall effectiveness of
the method in the evaluation described in Section \ref{sec:evaluation}.

%% file: 05_sentences.tex
\section{Similarity Function Tuning and Performance}
\label{sec:sim_function_tune}

In order to facilitate fine-tuning and evaluation
of the interlingual text passage similarity function presented in Section
\ref{sec:alignment_function}, we created a benchmark $Sim-B$. 
This benchmark defines the similarity scores for the interlingual text
passage pairs in the German and the English languages based on shared semantic
information.
We use parts of this benchmark for the fine-tuning of the similarity 
function as well as for the evaluation as discussed in the following.

\subsection{The \textit{Sim -- B} Benchmark for
Interlingual Similarity Computation}
\label{sec:sim-b-benchmark}

An important question for the benchmark creation is the selection of the text
passage pairs to be annotated. On the one hand, the annotation of all possible
text passage pairs in the partner articles does not appear feasible due to their
large number.
On the other hand, the
majority of text passage pairs that can be built in any partner 
article pair is rather dissimilar.
Therefore, random selection of text passages would not lead to a sufficient
number of similar pairs to train the similarity function. 
Hence, in order to
fine-tune the feature weights $\beta_{f_i}$ in Equation \ref{eq:al_function}, we apply an 
iterative bootstrapping approach. This approach incrementally collects
relevant text passage pairs and systematically refines the weights using
supervised machine leaning and user feedback.
For simplicity of the annotation, in $Sim-B$
we focus on short text passages, each consisting of a single sentence. 
In particular, we create three datasets:

\textbf{The dataset $TD_{1}$:} We first pre-select sentence pairs from a
set of controversial Wikipedia articles \cite{Yasseri2014} in the German and the
English languages aligned via the language links to build the training dataset
$TD_{1}$.
This is performed by using the text passage similarity function
following Equation \ref{eq:al_function} with a set of initial
manually defined feature weights. In order to include the sentences
that do not contain all features, we varied the feature weights, including
$0$-weights for the features based on the entity and time annotations.
These sentence pairs are judged by users, such that we can learn feature
weights for the similarity function using supervised machine
learning (in particular, we utilize linear regression for this task).

\textbf{The dataset $TD_{2}$:} Then we iteratively refine the feature
weights and incrementally collect sentence pairs for the second training
dataset $TD_{2}$. This dataset contains randomly selected partner articles.
We collect sentence pairs having similarity above a manually defined
threshold ($0.25$) and further refine the feature
weights.
When no substantial changes in the feature weights are observed in the next
iterations, we consider the feature weights to be optimal.

\textbf{The dataset $VD$:} Finally, we create a validation dataset $VD$. We
use this dataset to evaluate the similarity function. This
dataset contains sentence pairs extracted from randomly
selected partner articles using pooling  -- a standard evaluation method 
in Information Retrieval \cite{manning2008}.
To this extent, 
we retrieve the ranked list of the most similar sentence pairs generated by
different similarity functions (the functions are described in Section
\ref{sec:performance}). We ensure that the top-$k$ results of each function are user-annotated.
Other sentence pairs, i.e. those ranked below $k=200$ by all similarity
functions are considered to be dissimilar. 
Pooling method enables a fair comparison of the precision and recall values
across the functions considered in the evaluation, even though the absolute recall
values can be overestimated.

Table \ref{tab:datasets} provides an overview of the article selection
method and the size of the datasets.

\begin{table}[h]
	\centering
	\small
	\setlength{\tabcolsep}{4pt}
	\begin{adjustbox}{width=1\textwidth}
	\begin{tabular}{|l||l|r|r|r|}
		\hline
		\textbf{Dataset} & \textbf{Source}        &  \multicolumn{1}{c|}
		{\textbf{Articles}}
		& \multicolumn{1}{c|}
		{\textbf{\begin{tabular}[c]{@{}c@{}}Possible\\ Sentence Pairs\end{tabular}}} &
		\multicolumn{1}{c|}{\textbf{\begin{tabular}[c]{@{}c@{}}Annotated\\ Sentence Pairs\end{tabular}}} \\ \hline \hline
		\textbf{$TD_{1}$} & Controversial Articles & 14
		&
		2016568 & 229                                                                                                        \\
		\hline \textbf{$TD_{2}$} & Random Articles        & 20
		& 260867 & 1233
		\\ \hline \textbf{$VD$}  & Random Articles        & 33
		& 20358 & 300                                                                                                \\
		\hline
	\end{tabular}
		\end{adjustbox}
	\caption{Datasets consisting of the German and the English Wikipedia
	partner articles.
	During the function tuning and evaluation process,
	a subset of highly ranked sentence pairs aligned by different methods
	in each of these datasets has been annotated.}
	\label{tab:datasets}

\end{table}

\subsection{Similarity Annotations with Users}
\label{sec:sim_benchmark}

During the benchmark creation process described in Section
\ref{sec:sim-b-benchmark}, we annotated the initial training dataset $TD_{1}$ with $258$ pre-selected
sentence pairs in a user study.
In total, 11 users (graduate CS students with good knowledge of both languages)
participated in the user study.
Each user performed at least 50 tasks (an average evaluation time of a set of 50 tasks
was 30 minutes).
In each task, the user was presented the English
sentence and a list of one or more alignment candidates in German. The users
were asked to classify
each candidate as one of: ``same content (i.e. facts)'', ``partly same content'' or
``different content'' categories. In addition we made the options ``don't know''
and ``corrupted sentence''
available to the users. The last option helped to exclude sentences
containing occasional errors introduced
by the pre-processing from the evaluation.
Each sentence pair in this dataset was evaluated by at least 8 users.
In addition, we created a set of 12 manually selected
parallel sentence pairs as well as 12 randomly selected mismatched sentence
pairs to verify the user's input.

To compute the overall user-defined similarity scores for each sentence pair
in this training dataset, we
assigned the scores of 1.0 to the ``same content'', 0.5 to the ``partly same content'' and
0.0 to the ``different content'' judgements and
computed the similarity of a sentence pair as an average of the user scores.
As a result of the user study, we obtained an initial training dataset containing
229 aligned sentence pairs
(29 corrupted sentences and the sentence pairs added for verification
of users' input are ignored): 18 pairs with an average user score in the interval $[0.75,1]$ (\textit{parallel sentence pairs}),
102 pairs in $(0.25,0.75)$ (\textit{partially similar}) and 109 pairs in $[0,0.25]$ (\textit{different}). According to these numbers,
there is at least a
partial overlap in more than the half of the evaluated sentence pairs.

The datasets $TD_{2}$ and $VD$ have been
annotated using the same procedure, while employing a smaller number of
annotators. Table \ref{tab:simb_examples} shows two examples of 
the sentence pairs with their ratings in the $Sim-B$ benchmark.

\begin{table}[]
	\setlength\extrarowheight{2pt}
	\setlength{\tabcolsep}{4pt}
	\scriptsize
	\centering
	\begin{tabular}{|P{1.4cm}|l|l|l|P{2.8cm}|P{2.8cm}|}
		\hline
		\multicolumn{1}{|c|}{\textbf{Page}} & \multicolumn{1}{c|}{\textbf{\begin{tabular}[c]{@{}c@{}}Avg.\\ Rating\end{tabular}}} & \multicolumn{1}{c|}{\textbf{Article ID$_1$}} &  \multicolumn{1}{c|}{\textbf{Article ID$_2$}} & \multicolumn{1}{c|}{\textbf{Text$_1$}} & \multicolumn{1}{c|}{\textbf{Text$_2$}} \\ \hline
		European Union & $1.0$ & en-635761078 & de-136109478 & In 2012, the EU was awarded the Nobel Peace Prize. & 2012 wurde der Europäischen Union der Friedensnobelpreis zuerkannt. \\ \hline
		Nicolaus Copernicus & $0.4375$ & en-634443003 & de-134393612 & He died about 1483. & Als sein Vater 1483 starb, war Nikolaus zehn Jahre alt. \\ \hline
	\end{tabular}
	\caption{Example text passages from the $Sim-B$ benchmark.
	In addition to the sentence pairs and similarity 
	user ratings illustrated in this
	table, the benchmark also contains additional 
	(dataset) IDs, the single user ratings, the articles, 
	their sentences, semantic annotations and machine translations.}
	\label{tab:simb_examples}
\end{table}

\textbf{Difficulty of the similarity annotation task:}
To obtain a better understanding of the task difficulty for the
users, we computed the Fleiss' $\kappa$ \cite{kappa}, 
a statistical measure of agreement between
individuals for qualitative ratings, as a measure of the reliability of the
user agreement.
Note that according to Fleiss' definition, $\kappa < 0 $  corresponds to no agreement,
$\kappa =0 $ to agreement by chance, and $0 < \kappa \leq 1 $ to agreement beyond
chance.
Here, we considered the seven users with the most ratings.
Each of these users evaluated the majority of the 229 sentence pairs in 
$TD_{1}$.
If we do not differentiate between the partially overlapping
and the parallel sentence pairs, $\kappa$ value reaches $0.571$, which is
close to the ``substantial agreement''.
For the three classes (``same content'', ``partly same content'' and
``different content''), agreement values are lower and correspond to 
a ``moderate agreement'' by using the intervals
presented in \cite{Landis:1977} ($\kappa \approx 0.474$). In other terms,
we could find that for 131 sentence pairs (57.21\%), there has been at most 1
user disagreeing with the other users.
According to these values, it is presumably easier for the users to decide
whether two sentences are at least partially overlapping, than to 
differentiate between partially overlapping and parallel sentences in this
corpus. 

\textbf{Sources of disagreement by similarity annotations:}
To better understand the reasons for the users' disagreement, we looked at those
sentence pairs that were put into different classes by the users.
We found that user disagreement can be typically observed in cases with differences in the
author perspective and with missing context:

\begin{itemize}
  \itemsep0pt
\item Difference in the author perspective, generalization:

\begin{itemize}
\item English: ``Berlin is known for its numerous cultural institutions, many of which
enjoy international reputation''.\footnote{\url{http://en.wikipedia.org/wiki/Berlin?oldid=635429067}}
\item German: ``Die Sportereignisse, Universit\"aten, Forschungseinrichtungen und Museen
Berlins genie{\ss}en internationalen
Ruf''.\footnote{\url{http://de.wikipedia.org/wiki/Berlin?oldid=136234983}}
(Translated: ``The sport events, universities, research institutions and museums of Berlin
enjoy international reputation''.)
\end{itemize}

\item Missing context: References to other sentences, where the user has to consider the
context of the sentence in the Wikipedia article to disambiguate the reference:
\begin{itemize}
\item English: ``The church was destroyed in the Second World War and left in ruins''.
\item German: ``Sie war durch Bombenangriffe im Zweiten Weltkrieg schwer besch\"adigt worden''.
(Translated: ``It was heavily damaged by bombings in the Second World War''.)
\end{itemize}

\end{itemize}

Although both sentence pairs in these examples contain similar information and could be
viewed as parallel, some users classified them as partially overlapping
or even different due to the language-specific differences in the information presentation.
In order to increase the user agreement in the second case, missing
context could be provided
by presenting larger text passage context (e.g. paragraphs) to the users.

\textbf{Usage of the user annotations for function training and evaluation:} 
Although the absolute scores provided by the individual users can vary for some text passage pairs, 
the scores aggregated over
multiple user judgments provide a comprehensive picture of the relative
similarity across the text passage pairs. 
Such aggregated scores can be effectively used for 
training and evaluation of the similarity function.

\subsection{Similarity Function Evaluation}
\label{sec:performance}

In this work we establish the baseline for the
alignment of text passages
containing overlapping information nuggets across languages. 
Closest to our work, \citeN{Duh:2013} aimed at the identification of
new information in Wikipedia articles and applied cosine similarity measure.  
This method uses terms obtained after machine translation, stop word removal and
stemming.
    We use cosine similarity in isolation
    as a baseline to assess the relevance of the other semantic features
    we proposed.
We use the following notations for the similarity function configurations:

\begin{itemize}
  \itemsep-1pt
    \item \textbf{Co}: Cosine similarity. This method uses
    terms obtained using machine translation, stop word removal and stemming.
 \item \textbf{CoE}: This function combines Cosine similarity with
\textit{Entity Annotations}.
\item \textbf{CoT}: This function combines Cosine similarity with \textit{Time
Annotations}.
\item \textbf{CoET}: This function combines Cosine similarity with both,
\textit{Entity Annotations} and \textit{Time
Annotations}.
\end{itemize}

For the validation dataset $VD$, 
we collected the top-$200$ sentence pairs per similarity function under consideration. Given the resulting set of sentence pairs and the average user ratings per sentence pair, we compute the average precision ($AP$) values of the rankings achieved by the different alignment functions. For each sentence pair, its user similarity score is computed as an average over the scores given by the individual annotators. To determine the relevance of the retrieved sentence pairs for the average precision computation, we apply a threshold of $0.25$ on the average user ratings (i.e. we assume that the sentence pair is relevant if an average user rating is $\geq 0.25$).

\begin{table}[]
	\centering
	\small
\begin{tabular}{|c||r|r|}
	\hline
	\multirow{3}{*}{\textbf{\begin{tabular}[c]{@{}c@{}}Similarity\\ Function\end{tabular}}} & \multicolumn{2}{c|}{\textbf{Average Precision (AP)}}                                                                                                                                   \\
	& \textbf{\begin{tabular}[c]{@{}c@{}}All\\ Sentences\end{tabular}} &
	\multicolumn{1}{c|}{\textbf{\begin{tabular}[c]{@{}c@{}}Without Parallel\\
	Sentences\end{tabular}}} \\ \hline \hline
		\multicolumn{1}{|l||}{\textbf{Co}}        & \multicolumn{1}{|l|}{84.35\%}                     & 74.99\%                                           \\ \hline
		\multicolumn{1}{|l||}{\textbf{CoE}}       & \multicolumn{1}{|l|}{85.73\%}      
		& 76.79\%                                    \\ \hline
		\multicolumn{1}{|l||}{\textbf{CoT}}       & \multicolumn{1}{|l|}{89.77\%}                     & 83.32\%                                    \\ \hline
		\multicolumn{1}{|l||}{\textbf{CoET}}      &
		\multicolumn{1}{|l|}{\textbf{90.98\%}}                     & \textbf{84.63\%}                                    \\ \hline
	\end{tabular}
	\caption{Average precision values for text passage similarity functions in the
	alignment task.
	For each of the functions, the top-$200$ sentence pairs were collected.
	Based on the resulting set of $267$ sentence pairs 
	($257$ for non-parallel sentence pairs), 
	the average precision values were computed.}
	\label{tab:pr_diagram_overlap}
\end{table}

As we can observe in Table \ref{tab:pr_diagram_overlap}, 
\textbf{Co} that only uses \textit{Cosine Similarity} achieves an
average precision of $84.35\%$.
Additional semantic features we proposed in this article 
such as \textit{Entity Annotations} in \textbf{CoE} ($AP=85.73\%$) and
\textit{Time Annotations} in \textbf{CoT} ($AP=89.77\%$) 
and in particular the combination of these annotations in \textbf{CoET}
($AP=90.98\%$) enable us to further improve the average precision. 
This result confirms the high effectiveness of the proposed 
semantic features for the text passage alignment.

When only considering partially overlapping sentences, 
several differences can be observed: 
The absolute average precision values of all similarity functions drop,
confirming that it is easier to align the parallel sentences than the partially
overlapping ones. This decrease varies dependent on the similarity function:
Compared to the case with the parallel sentences, \textbf{CoET} shows the lowest 
decrease, which emphasizes the value of the semantic annotations for the
alignment of partially overlapping sentences: In this case \textbf{CoET}
achieves up to 9.64\% improvement in the average precision compared to the purely 
syntactic-based function \textbf{Co}.

An improvement when using the \textit{Time Annotations} is higher 
than for the \textit{Entity Annotations}, because although very selective 
\textit{Entity Annotations} contribute to the precise text passage alignment, 
we cannot use less selective annotations
to precisely differentiate the sentences with 
semantically meaningful information nugget overlap from the rest.

Overall, our evaluation results confirm that the use of semantic features leads to 
a more precise text passage alignment, in particular with respect to the
partially overlapping sentences and indicates that \textit{Time Annotations}
and selective \textit{Entity Annotations} are effective features towards
identifying semantically similar text passages containing common information
nuggets.

%% file: 06_userstudy.tex
\section{Text Passage Alignment Benchmark}
\label{sec:userstudy}

In order to better understand the problem of the interlingual text passage
extraction and alignment from the user perspective we
collect the benchmark $Align-B$ for the method tuning and evaluation in a user
study.
The aims of the user study were to: 1)
Better understand the difficulty of the manual text passage alignment task for the users; 2)
Observe, analyze and learn from the user decisions regarding the alignment of
text passages; and to 3) Create a benchmark to fine-tune and evaluate automatic methods for this
task.

\subsection{The $Align-B$ Benchmark for Text Passage Alignment}
\label{sec:align-b-benchmark}

To collect the user annotations for the interlingual text passage alignment, 
we randomly selected a set of partner articles 
from the English and the German Wikipedia. As we focus on the text passages, 
article pairs where one of the articles mainly consisted of tables and
lists were filtered out manually. The resulting dataset contains $55$ article pairs
coming from several domains and includes, for example, 
``General Post Office'', ``Commuter Rail'' 
and ``George William Gray''.
With regard to 
the split obtained by a sentence splitting algorithm, 
the English articles contain 21.32 sentences on average and the 
German ones 17.27, which makes a total of 2,123 sentences.
Based on the $55$ article pairs in this dataset, we created another
dataset from the Russian and English Wikipedia. This dataset consists 
of the $21$ article pairs whose articles can
be found in the Russian Wikipedia as well.

In the user study, the user's task was, given a pair of German-English or Russian-English
partner articles, to extract and align similar text passages. 
In total, 12 users (graduate CS or mathematics students with good knowledge of both languages)
participated in the user study for the German-English article pairs. Each user annotated 15 article pairs on
average.
All 55 article pairs were annotated by at least three users each, 
14 of them by four different users. 
This makes a total set of 179 distinct user annotations of article pairs. 
In each of these annotations, at least one text passage 
pair was identified. The average number of text passage pairs annotated 
per article pair is 3.34. 
The 21 articles in the Russian-English dataset were
manually annotated by at least one user.
Three example text passage pairs
in $Align-B$ are shown in Table \ref{tab:alignb_examples}.

\begin{table}[]
	\setlength\extrarowheight{2pt}
	\setlength{\tabcolsep}{4pt}
	\scriptsize
	\centering
	\begin{tabular}{|l|l|l|l|l|l|l|}
		\hline
		\multicolumn{1}{|c|}{\textbf{Page}} & \multicolumn{1}{c|}{\textbf{User ID}} & \multicolumn{1}{c|}{\textbf{Article ID$_1$}} &  \multicolumn{1}{c|}{\textbf{Article ID$_2$}} & \multicolumn{1}{c|}{\textbf{Passage$_1$}} & \multicolumn{1}{c|}{\textbf{Passage$_2$}} & \multicolumn{1}{c|}{\textbf{Title}} \\ \hline
		\begin{tabular}[c]{@{}l@{}}Winger\\ (sports)\end{tabular} & 43 & \scriptsize{en-664310306} & \scriptsize{de-664310306} & \begin{tabular}[c]{@{}l@{}}10-11\\ -12-13\end{tabular} & 7-8-9 & Football \\ \hline
		\begin{tabular}[c]{@{}l@{}}Johann Hugo\\ von Orsbeck\end{tabular} & 43 & \scriptsize{en-668382450} &  \scriptsize{de-144384256} & 4-5 & 8-9 & \begin{tabular}[c]{@{}l@{}}Birth and\\ parents\end{tabular} \\ \hline
		\begin{tabular}[c]{@{}l@{}}Johann Hugo\\ von Orsbeck\end{tabular} & 47 & \scriptsize{en-668382450} & \scriptsize{de-144384256} & 29 & 61 & The end \\ \hline
	\end{tabular}
	\caption{Example text passages from the $Align-B$ benchmark.
	In addition to the user-aligned text passage pairs illustrated in this table,
	the benchmark also contains the actual sentences.}
	\label{tab:alignb_examples}
\end{table}

\subsection{Task Description and User Interface}
\label{sec:UserInterfaceStudy}

The user interface of the study is similar to the interface 
shown in Fig. \ref{fig:introduction_example}: At the beginning, the user sees 
both articles without any marked text passage pairs. 
By clicking on the sentences, the user can incrementally create text passages in
both articles simultaneously, expand the text passages by adding further 
sentences on each language's side and finally confirm the alignment of the
created text passage pair.
To ensure topical coherence of the created text passage pair, 
the last step requires an input of a brief user-defined English title. 

The instructions for the user included the following steps: 
\begin{enumerate}
	\item [Step 1] Read both language versions of the article.
	\item [Step 2] Find and align a pair of similar text passages 
	in the two articles. 
	If several alignment candidates are available, select only the best matching pair. 
	If similarity and topics are the same, prefer longer text passages.	
	\item [Step 3] Give the aligned text passage pair an English title.
	\item [Step 4] Continue with Step 2 until all similar text passage pairs 
	across both language editions of the article are aligned. 
\end{enumerate}

With these instructions we let the decision if 
any intermediate sentences should be included to the user, 
as long as the aligned text passages fulfill the conditions
specified in Step 2.
 
\subsection{Difficulty of the Text Passage Alignment Task}
\label{sec:difficulty-alignment}

In order to better understand the difficulty of the task for the users, 
we measured the time spent by the
users on the task and the user agreement. 
On average, a user spent approx. 6 minutes to annotate one article pair. The 
annotation time depends on the article length: If one of the articles is very
short, users spent less than a minute; On longer articles, some users spent 
more than 15 minutes.
Overall, we can observe that the task of text passage alignment can be
very time consuming especially for longer articles, even for users with good knowledge 
of both languages.

To capture the overlap across text passages aligned by different users, 
we consider the intra- and interlingual sentence pairs within the annotated articles 
and check how different users assign 
these sentence pairs to text passages. 
In particular, the intra-lingual measurement estimates the agreement of the users on 
the extraction step
(i.e. for each article we create the set of all possible sentence pairs and check for
each sentence pair and rater whether the two sentences were put in the same text 
passage);
The interlingual alignment step
illustrates the agreement of the users on the alignment step (i.e. the agreement 
that two sentences, one 
from each partner article, belong to an aligned interlingual text passage pair). 

We again compute the inter-rater agreement using Fleiss' $\kappa$-measure.
When considering short articles (less than 2500 characters in both articles together), 
$\kappa$-values 
for both the extraction task and the interlingual alignment task
approach $0.6$ indicating substantial agreement. 
Considering all article pairs independent of their length, 
we measured $\kappa \approx 0.467$ for the extraction task, and
$\kappa \approx 0.473$ for the interlingual alignment task.
These $\kappa$ values illustrate that the extraction and alignment tasks are 
similar with respect to their 
difficulty and moderate agreement is possible even in case of longer articles.
This also confirms our initial observation performed by the time measurement, 
that in case of longer articles the task becomes 
more difficult for the users.

\subsection{Observations}
By analyzing the user annotations we made important observations 
with respect to the annotation structure 
that can help us to further fine-tune the 
text passage alignment model.

\textbf{Alignment of the lead sentences:} 
The lead sentence, i.e. the very first sentence of a Wikipedia
article, and the lead sections of Wikipedia articles typically have a uniform
style:
Regarding the English Wikipedia manual of style, an article lead 
section ``should be able to stand alone as a concise overview'' and the first 
sentence ``should tell the non-specialist reader 
what (or who) the subject
is.''\footnote{\url{https://en.wikipedia.org/wiki/Wikipedia:Manual_of_Style/Lead_section}}
This subject description is rather independent of the specific language and thus it
is very likely that there is an aligned text passage pair containing at least
the lead sentence of both partner articles. In our study, this was the case for
$92.74\%$ of all the article pairs.

\textbf{Comparison of the text passage length:} 
As we enable users to include additional 
information in the aligned text passage pairs, although it only occurs 
in one of the text passages as long as it is related to a common topic, 
the length of the aligned text passages can differ. The results of our 
user study show that only $51.59\%$ of the user-aligned text 
passage pairs have exactly the same number of sentences in both languages. 
On average, the length of the aligned text passages differs by $0.89$ sentences; 
Overall, the interlingual difference in the length of the user-aligned text 
passages follows a power law 
distribution and can exceed three or more sentences in $8.21\%$ of the cases.

\textbf{Text passages vs. text paragraphs:} 
We assume that each text paragraph and section in the original Wikipedia
text hierarchy can form a topically coherent text passage. 
To estimate the usefulness of the Wikipedia text structure in the context
of text passage alignment, we measured how the
user-extracted text passages correlate with the article structure. 
We observed that out of $642$ text passages extracted by the users (excluding
those consisting of one sentence only), $481$ ($74.92\%$) are entirely contained within a 
single Wikipedia text paragraph, with $241$ ($37.54\%$) of them even being
equivalent to the Wikipedia text paragraph.
$626$ ($97.51\%$) text passages are placed within the same Wikipedia section.
Although few text passages span across Wikipedia sections, 
such extraction is less typical.

\textbf{Text passage titles:} For each aligned text passage pair, the users provided titles. 
These titles can be roughly assigned into three categories as follows:

\begin{itemize}
	\item Lead section of the article: ``Summary'', ``Description'', ``Definition'', ``Name'', ``Short biography''.
	\item Typical sections: ``History'', ``Family'', ``Career'', ``Early life'', ``Demographics'', ``Awards''.
	\item Article-specific: ``Prime minister'', ``Ice hockey'', ``The accident'', ``Bishop of Speyer and Trier''.
\end{itemize}

We manually
categorized the titles given by the users and found that 130 titles (22\%)
belong to the first, 143 titles (24\%) to the second and 324 titles (54\%) to the last category. 
The first category's titles reoccur very often (``Summary'', ''Description'' and ''Definition'' are the three most frequent titles) 
and illustrate the aforementioned observation that the lead paragraphs in each article are likely to be aligned. 
The titles in the first two categories come from a rather small set of titles and depend on the entity type 
(e.g. the articles about countries and cities often contain information about demographics). 
In the last category, there are very specific titles that may be unique for an article and often are more detailed 
than Wikipedia section titles. 
In this article we
do not perform any automatic labeling of the aligned text passages to annotate
the topics. These observations can help to perform automatic labeling in
future research.

\textbf{Disagreement sources:} 
Typical sources of annotator disagreement in our dataset 
include the differences in the granularity of extracted topics 
and the varying level of details across the partner articles. 

\textit{Granularity of extracted topics:} 
The annotators can disagree on the granularity of the topics and the 
corresponding text passages to be extracted.
For example, there is a section
about the political career of the Lithuanian politician Algirdas Butkevi\v{c}ius with about 
ten sentences in each the English and the German articles. 
One of the annotators performed an alignment of the complete section as a
single text passage and entitled it as ``Political carrier''. Another annotator
performed a higher granularity alignment by splitting 
it into two text passages entitled ``Begin of political carrier'' and ``Prime
minister''; the third annotator created three text passages: 
``SDLP\footnote{Social Democratic Party of Lithuania} 
Membership'', ``Minister of Finance'' 
and ``Prime Minister''. 

\textit{Level of description details:} Another source of disagreement is the
case of the interlingual differences between the articles where a very specific description in one article
corresponds to a more generic and less detailed description in the partner
article.
For example, in the English article about the European pine vole, 
there is only a list of the countries where 
the animal lives, while the German article has a whole section with a well-phrased text 
about the animal's habitat. 
Only one out of three users performed the alignment between the country list
and the detailed description.

As certain disagreement is expected in such cases, we treat all
user annotations as correct alternatives during the evaluation.

%% file: 07_paragraphs_evaluation.tex
\section{Evaluation of Interlingual Text Passage Alignment}
\label{sec:evaluation}

To the best of our knowledge, the problem of interlingual text passage alignment
presented in this article has not been addressed by any existing approach. 
To enable an evaluation of the proposed method, we use state-of-the-art 
methods for text segmentation and alignment 
that take different approaches to the extraction and alignment
aspects as baselines.
We evaluate our approach 
using the user annotated $Align-B$ benchmark presented in
Section \ref{sec:userstudy} and experimentally demonstrate the effectiveness of
our method and its superiority with respect to the baselines.

\subsection{Methods and Baselines}
\label{subsec:alignment_functions}

To enable effective interlingual text passage alignment,
\textit{MultiWiki} relies on two main components: 1) Extraction of text
passages taking their interlingual context into account; and 2)
Interlingual alignment of the extracted text passages. In order to evaluate
\textit{MultiWiki}, we use baseline methods, each taking a different approach on the
extraction and alignment steps:

\begin{itemize}
	
	\item \textbf{Sentence alignment baseline (\textit{SA Baseline}):} 
The \textit{SA Baseline} aligns interlingual sentence pairs using 
	a state-of-the-art sentence alignment function defined in \cite{Duh:2013}.
	In contrast to our method, the \textit{SA Baseline} does not merge aligned 
	sentences into longer text passages. Therefore, a comparison between our method and this baseline can highlight the impact  
	of the text passage extraction on the evaluation metrics.
	As the sentences pairs aligned by this baseline are syntactically similar, we expect \textit{SA Baseline} 
	to achieve high precision, but at the price 
	of a significant granularity increase compared to our method.
	
	\item \textbf{Plagiarism detection baseline (\textit{PD Baseline}):} 
	The goal of the plagiarism detection is to identify contiguous maximal-length 
	passages containing reused text \cite{Perez:2015}.
	While our problem is more general and includes 
	a broader range of semantically similar text passages, topically coherent 
	plagiarism text passages can constitute valid  
	alignments according to our definition.
	To facilitate a comparison, we use a state-of-the-art plagiarism 
	detection method as a baseline \cite{Perez:2015}. 
	As plagiarism detection can be viewed as a special case of text passage alignment, 
	we expect this baseline to achieve lower recall compared to our method.

	\item \textbf{Alignment of Wikipedia paragraphs (\textit{WikiParagraphs}):}
	The \textit{WikiParagraphs} method takes the text paragraphs as specified in the original 
	structure of the Wikipedia articles. We perform the alignment of such pre-defined paragraphs 
	equivalently to \textit{MultiWiki} using the same similarity function and thresholds.
	Compared to the method proposed in this article, in \textit{WikiParagraphs} the
	boundaries of the text passage are defined a-priori by the Wikipedia structure
	and do not take into account any interlingual context.
	As such paragraphs are user-defined, we expect
	\textit{WikiParagraphs} to perform well with respect to the granularity of the alignment, 
	as such paragraphs should be intuitive for human readers. 
	However, as \textit{WikiParagraphs} misses text passages deviating from the Wikipedia text 
	paragraphs, we expect to obtain lower recall values. In addition, the interlingual differences in 
	the paragraph structure can affect precision of the alignment. 
		
	\item \textbf{Alignment of TextTiling segments (\textit{TextTiling}):}
	This baseline represents the TextTiling algorithm \cite{Hearst:1997} that 
	subdivides texts into topically coherent segments using term distributions irrespective of the original text structure. 
	 We perform the interlingual alignment of such segments equivalently to the \textit{MultiWiki} and \textit{WikiParagraphs} methods.
	 Similarly to \textit{WikiParagraphs}, the segmentation performed by \textit{TextTiling} 
	  does not take the interlingual context of the article into account. 
	Given their comparable approaches, we expect the results of \textit{TextTiling} and \textit{WikiParagraphs}
	to be similar.  
\end{itemize}

As discussed in the problem statement in Section \ref{sec:problem}, 
the most important criterion for 
an effective text passage alignment is to achieve an optimal combination 
of the precision, recall and granularity of the aligned text passage pairs. 
Whereas the individual baselines are naturally optimized for one of these metrics, 
we expect our method to achieve the best performance with respect to their
combination.

\subsection{Dataset for Interlingual Text Passage Alignment}
\label{sec:dataset}

To facilitate the evaluation of the effectiveness of \textit{MultiWiki} 
as well as the parameter tuning, 
we randomly split the German-English part of the $Align-B$ benchmark
defined in Section \ref{sec:userstudy} into two disjoint parts.
The first part, used as a training dataset, contains $20$ article pairs. We
refer to this part as $Align-T$ and use it for the parameter
tuning of \textit{MultiWiki} presented in Section \ref{sec:tuning}. The other part, a
validation dataset $Align-V$, contains the remaining $35$ article pairs
and is used to evaluate our approach. The Russian-English subset of
the benchmark is named $Align-R$.

Our $Align-B$ benchmark includes article pairs annotated by up to four
different users.
These user annotations can indicate some differences with respect to the
text passage extraction and alignment.
It does not appear feasible to build a single alignment incorporating
all possible alignments of different users on the same article: even small 
deviations in the user annotations (e.g. one extra sentence added to a text
passage) would lead to an overall different annotation of the article that
cannot be directly merged into a single representation without modifying the
user-defined alignment. Therefore, in the evaluation, we first compute the
scores for each user and each metric separately and then aggregate 
them to build average scores over the users per article, normalized by the
article lengths.

\subsection{Evaluation Metrics}
\label{sec:metrics}

The goal of our evaluation is to compare the quality of the text passage alignment 
using  
the methods discussed before with respect to their precision, 
recall and granularity. Although we expect different methods to optimize 
some of these metrics in isolation, we are particularly interested in 
measuring their overall impact.  
According to the problem statement defined in Section \ref{sec:problem}, 
such evaluation measure must reward high similarity and penalize high 
granularity of the alignment. 
In addition, high recall is important to ensure 
that as many user-aligned text passage pairs are found as possible.

These requirements are to a large extent addressed by 
the \textit{plagdet} metric as defined in \cite{Potthast:2010}. 
In the following we discuss this metric and the adjustments we made to fit this 
metric (originally defined in the context of the plagiarism detection) our
problem, in particular with respect to the granularity computation. 

The \textit{plagdet} metric is based on the character-based precision 
and recall scores as well as on an additional score for granularity. 
Put together, the three measures form the \textit{plagdet} 
score that is used as an overall measure of the alignment effectiveness. 
\textit{Plagdet} requires a set $S$ of the user-defined text passage 
alignment cases (ground truth) and a set $R$ 
of their algorithmic detections. 
In our case, the set $S$ is obtained 
from the $Align-B$ benchmark.
Each method to be evaluated returns its own set $R$ of text passage detections.

In the context of the interlingual text passage alignment, 
an adjustment is required for the granularity computation.
Plagiarism detection is a directed problem: Given a suspicious document and a source 
document, the goal is to detect reused text parts in the suspicious document. 
Therefore, in the original \textit{plagdet} metric, the granularity is defined
as a measure of ``whether a plagiarism case $s \in S$ is detected 
as a whole or in several pieces'' \cite{Potthast:2010}. That means, it would be sufficient if the alignment 
algorithm would return longer text passages to perform well with respect to this measure. 
In our case we aim at matching text passages in both articles simultaneously, 
as close as possible to the user-defined extraction. Therefore, the measure should reflect the granularity 
on both sides of the alignment. Thus, we define the symmetric granularity measure 
$gran_{symm}(S,R)$ to be used in our computation of the \textit{plagdet} score:

\begin{equation}
gran_{symm}(S,R) = \frac{gran(S,R) + gran(R,S)}{2}.
\end{equation}

In more detail, the individual components of \textit{plagdet} are defined as
follows:

\begin{itemize}
	\item \textbf{Precision:} The fraction of the characters in $R$ that are among the 
	user-defined text passage pairs. 
	Precision computed at the character level is further normalized using the text passage lengths.
	\item \textbf{Recall:} The fraction of characters in $S$ that are determined by the algorithm.
	\item \textbf{Plagdet:} A combination of the previous scores that rewards a high $F$ value 
	(harmonic mean of precision and recall) and low 
	granularity: $plagdet(S,R) = \frac{F_1}{log_2(1+gran_{symm}(S,R))} \in [0,1]$.
\end{itemize}
	
\textbf{Granularity and Inverse Granularity:} 
	The $gran_{symm}$ measure defined above is anti-correlated with the effectiveness of the method 
	(with  $gran_{symm}=1$ being the best and $gran_{symm}=|R|$ being the worst).
	To simplify the presentation of the results we will also use $I-Gran(S,R)$ -- i.e. its inverse value:  
	$I-Gran (S,R) = gran_{symm}(S,R)^{-1} \in [\frac{1}{|R|},1]$. The $I-Gran$
	measure is positively correlated with the method effectiveness with respect to the granularity aspect.

\subsection{Parameter Tuning}
\label{sec:tuning}

To identify the optimal values for the threshold $th$ and
the structure freedom $sf$ parameters, we used the $Align-T$ 
dataset described in Section \ref{sec:dataset}.
The best performing values of the parameters on the $Align-T$ are: 
Similarity threshold $th=0.21$ and structure freedom $sf=min$ 
(i.e. the text passage should stay within one Wikipedia text paragraph).
Moreover, as we observed that the lead sentences of both articles were 
aligned in $92.74\%$ of 
the user aligned articles, we apply a lower similarity threshold (i.e. $th/2$) for
the seed text passage pair consisting of the lead sentence per article.

Fig. \ref{fig:eval_configurations} gives a more detailed overview 
of how \textit{MultiWiki} behaves for different 
values of the individual parameters. 
Obviously, the similarity threshold (Fig. \ref{fig:th}) has a strong impact 
on the alignment results. If this threshold is too high, recall decreases as 
not enough sentence pairs are identified at the beginning of the alignment procedure. 
The best \textit{plagdet} score for the comparison with the user corpus is
achieved when setting $th \approx 0.21$. For very low values of $th$, 
recall decreases again: 
This is because the alignment function initially aligns a large number of sentence pairs, 
which decreases the effectiveness of the further extraction steps.

With respect to the structure freedom parameter $sf$ (Fig. \ref{fig:sf}), we can 
observe that the best 
results with respect to the precision, granularity and \textit{plagdet} scores 
are achieved with the $sf=min$ settings, meaning that the extracted text passages
should be contained within one text paragraph. That is consistent 
with our observation that 
users do not tend to extract text passages exceeding Wikipedia text paragraphs or sections. 
The increase in recall value for $sf=max$ accounts for the other cases,
where users selected longer text passages.

Overall, we observe that the similarity threshold 
and the structure freedom parameters are 
effective to control the alignment results. These parameters allow to put
an emphasis on the selected measures such as recall and granularity of the
alignment.

\input{parameter_diagram}

\subsection{Evaluation Results}
\label{sec:ev_results}

In this section we first compare the length of text passages extracted by different methods 
to get insights into their granularity. 
Then we present the evaluation results of the text passage alignment effectiveness 
achieved by our method and the baselines presented in Section \ref{subsec:alignment_functions}.

\textbf{Text passage length comparison:}
Table \ref{tab:alignment_statistics} provides an overview of the
average number of text passage pairs per article alignment and the 
number of sentences per text passage pair using $Align-T $ and $Align-V$. 
An optimal alignment should be equivalent to 
the text passages in the user alignment. 
As we can observe, the \textit{SA Baseline} that aligns individual sentences 
contains just two sentences per text passage pair, one in each language. 
In contrast, the \textit{PD Baseline} that comes from the plagiarism detection domain 
forms very long text passage pairs sometimes spanning across several 
Wikipedia sections, containing over $19$ sentences on average. 
Our method \textit{MultiWiki} comes closest to the user 
alignment ($4.43$ sentences per text passage pair,  
vs. $5.00$ sentences in the user-defined
text passages), which is a good indicator of an appropriate granularity.

\begin{table}[ht]
	\tbl{Number of text passage pairs and sentences aligned by different methods.\label{tab:alignment_statistics}}{
	\centering
	\small
	\begin{tabular}{|l||c|c|}
		\hline
		& \textbf{\begin{tabular}[c]{@{}c@{}}\# Text Passage Pairs\\ per Article Pair\end{tabular}} & \textbf{\begin{tabular}[c]{@{}c@{}}\# Sentences per\\ Text Passage Pair\end{tabular}} \\ \hline \hline
		\rowcolor{black!10} \multicolumn{1}{|l||}{\textbf{User Average}} & \textbf{3.34} & \textbf{5.00} \\ \hline
		\multicolumn{1}{|l||}{\textbf{PD Baseline}} & 1.32 & 19.55 \\ \hline
		\multicolumn{1}{|l||}{\textbf{TextTiling}} & 2.16 & 7.04 \\ \hline
		\multicolumn{1}{|l||}{\textbf{WikiParagraphs}} & 2.96 & 6.64 \\ \hline
		\multicolumn{1}{|l||}{\textbf{MultiWiki}} & 4.50 &  \textbf{4.43} \\ \hline
		\multicolumn{1}{|l||}{\textbf{SA Baseline}} & 6.13 & 2.00 \\ \hline	\end{tabular}}
\end{table}

\textbf{Effectiveness of the alignment methods:}
The effectiveness results achieved by different methods applied on
$Align-V$ and $Align-R$ with respect to the precision, recall, granularity and \textit{plagdet} metrics
are shown in Fig. \ref{fig:eval_baselines}. 

Fig. \ref{fig:eval_baselines_precision} presents the precision scores for the five 
alignment methods applied on the German-English article pairs in $Align-V$.
\textit{SA Baseline} and \textit{PD Baseline} achieve precision 
values of $83.48\%$ and $91.26\%$, respectively. This is expected, as both of these baselines 
specifically focus on the syntactic similarity, either by selecting individual 
sentences (\textit{SA Baseline}), or by extracting plagiarism cases (\textit{PD Baseline}). 
\textit{TextTiling}, \textit{WikiParagraphs} and \textit{MultiWiki} allow for additional content within 
the text passages and thus show lower precision values. \textit{WikiParagraphs} that uses 
predefined Wikipedia paragraphs shows $70.4\%$ precision, which is $8\%$ above \textit{TextTiling}.
This number can be significantly improved
by enabling flexible extraction in \textit{MultiWiki}, increasing precision to $82.41\%$.

This flexibility in the text passage extraction enables \textit{MultiWiki} to outperform 
other methods with respect to the recall metric. The comparison of the recall values is presented in 
Fig. \ref{fig:eval_baselines_recall}. Our algorithm \textit{MultiWiki} achieves over $58\%$ recall 
and outperforms the second best method (\textit{WikiParagraphs}) by $2.6$ points for that measure. 
The \textit{TextTiling} and \textit{PD Baseline} are least flexible, resulting in 
low recall values of $50.46\%$ and $41.77\%$. In particular, 
\textit{PD Baseline} detects too few 
text passage pairs, whereas \textit{SA Baseline} does not include enough
sentences.

Regarding the $I-Gran$ values depicted in Fig. \ref{fig:eval_baselines_granularity}, 
\textit{MultiWiki} ($I-Gran=0.86$), \textit{TextTiling} ($I-Gran=0.83$) and 
\textit{WikiParagraphs} ($I-Gran=0.90$) 
significantly outperform the baseline methods \textit{SA Baseline} ($I-Gran=0.73$)
and \textit{PD Baseline}  ($I-Gran=0.59$). 
These results confirm the observations made in Table \ref{tab:alignment_statistics} 
where these methods came closest to the user alignment. 
As \textit{WikiParagraphs} constitutes longer text passages, its $I-Gran$ scores are higher.

The \textit{plagdet} metric in Fig. \ref{fig:eval_baselines_plagdet} aggregates
the results of precision, recall and granularity. According to this metric, our \textit{MultiWiki}
method performs best and achieves the highest \textit{plagdet} score of $0.56$, 
that is $0.03$ points better than \textit{WikiParagraphs}. \textit{MultiWiki} 
outperforms the \textit{SA Baseline} by $0.09$ and \textit{PD Baseline} by $0.26$ points.
The results of the paired t-test confirm statistical significance of this
result for the confidence level of 95\%.

Overall, our evaluation results confirm the high effectiveness of our \textit{MultiWiki} method. 
\textit{MultiWiki} achieves the highest \textit{plagdet} and recall scores and  
outperforms the baselines with respect to granularity
due to its flexibility in the extraction process. 
This result also demonstrates that 
existing approaches like \textit{SA Baseline} and \textit{PD Baseline} that optimize 
for syntactic similarity cannot be effectively applied to the  
problem of interlingual text passage alignment presented in this
article.
When using the predefined paragraphs for the alignment, \textit{WikiParagraphs}
outperforms \textit{TextTiling} in every aspect. Hence, the text paragraphs in Wikipedia represent
a more intuitive division of the article into its subtopics from the user
perspective than the \textit{TextTiling} method.

\textbf{English-Russian dataset:} 
To confirm the generalizability of our approach on other language pairs, 
we evaluated the alignment methods on 
the Russian-English article pairs in $Align-R$, using the same values for the 
similarity functions weights and passage alignment parameters as in the English-German case. 
The results follow a similar distribution compared to the English-German 
evaluation. \textit{MultiWiki} achieves a \textit{plagdet} score of $0.63$, 
outperforming the other methods as seen in Fig. \ref{fig:eval_baselines_plagdet_ru}.
As we did not perform any language-specific training on the Russian-English
data, these evaluation results suggest that 
the training results obtained in one language can be effectively applicable to
the sentence similarity computation and text passage alignment in other
language pairs.

\begin{figure}[htbp]
	\centering
	\rotatebox[origin=c]{90}{German-English}\quad
	\begin{subfigure}{.23\linewidth}
		\centering
		\huge
		\resizebox{\linewidth}{!}{
			\begin{tikzpicture}[scale=1]
			\begin{axis}[
			ymax=1,ymin=0,bar width=20,enlarge x limits=0.25,
			symbolic x coords={M,W,T,S,P},
			xtick=data,xtick pos=left,ymajorgrids=true,
			ytick pos=left]
			\addplot[ybar,fill=blue!40] coordinates {
(M,0.824095516587727)
(W,0.7040073945486915)
(S,0.8347791420852054)
(T,0.622057253415823)
(P,0.9125674393317773)
			};
			\end{axis}
			\end{tikzpicture}
		}
		\caption{Precision}
		\label{fig:eval_baselines_precision}
	\end{subfigure} 
	\begin{subfigure}{.23\linewidth}
		\centering
		\huge
		\resizebox{\linewidth}{!}{
			\begin{tikzpicture}[scale=1]
			\begin{axis}[
			ymax=1,ymin=0,bar width=20,enlarge x limits=0.25,
			symbolic x coords={M,W,T,S,P},
			xtick=data,xtick pos=left,ymajorgrids=true,
			ytick pos=left]
			\addplot[ybar,fill=blue!40] coordinates {
(M,0.5896004592964412)
(W,0.5636750453434266)
(S,0.5544579616783677)
(T,0.5046119423383025)
(P,0.4177131553763895)
			};
			\end{axis}
			\end{tikzpicture}
		}
		\caption{Recall}
		\label{fig:eval_baselines_recall}
	\end{subfigure}
	\begin{subfigure}{.23\linewidth}
		\centering
		\huge
		\resizebox{\linewidth}{!}{
			\begin{tikzpicture}[scale=1]
			\begin{axis}[
			ymax=1,ymin=0,bar width=20,enlarge x limits=0.25,
			symbolic x coords={M,W,T,S,P},
			xtick=data,xtick pos=left,ymajorgrids=true,
			ytick pos=left]
			\addplot[ybar,fill=blue!40] coordinates {
(M,0.8586707876819446)
(W,0.8951532535765779)
(S,0.7253300958694596)
(T,0.828174941032328)
(P,0.5929504798715577)
			};
			\end{axis}
			\end{tikzpicture}
		}
		\caption{I--Gran}
		\label{fig:eval_baselines_granularity}
	\end{subfigure}
	\begin{subfigure}{.23\linewidth}
		\centering
		\huge
		\resizebox{\linewidth}{!}{
			\begin{tikzpicture}[scale=1]
			\begin{axis}[
			ymax=1,ymin=0,bar width=20,enlarge x limits=0.25,
			symbolic x coords={M,W,T,S,P},
			xtick=data,xtick pos=left,ymajorgrids=true,
			ytick pos=left]
			\addplot[ybar,fill=blue!40] coordinates {
(M,0.5612645470212733)
(W,0.5278380688505845)
(S,0.4681508005006123)
(T,0.4275473345276701)
(P,0.30266281502191406)
			};
			\end{axis}
			\end{tikzpicture}
		}
		\caption{Plagdet}
		\label{fig:eval_baselines_plagdet}
	\end{subfigure}
	\rotatebox[origin=c]{90}{Russian-English}\quad
	\begin{subfigure}{.23\linewidth}
		\centering
		\huge
		\resizebox{\linewidth}{!}{
			\begin{tikzpicture}[scale=1]
			\begin{axis}[
			ymax=1,ymin=0,bar width=20,enlarge x limits=0.25,
			symbolic x coords={M,W,T,S,P},
			xtick=data,xtick pos=left,ymajorgrids=true,
			ytick pos=left]
			\addplot[ybar,fill=blue!40] coordinates {
(M,0.8515571472714734)
(W,0.7414695314129544)
(S,0.8706565727484066)
(T,0.7492869679690163)
(P,0.8973124542484487)
			};
			\end{axis}
			\end{tikzpicture}
		}
		\caption{Precision}
		\label{fig:eval_baselines_precision_ru}
	\end{subfigure} 
	\begin{subfigure}{.23\linewidth}
		\centering
		\huge
		\resizebox{\linewidth}{!}{
			\begin{tikzpicture}[scale=1]
			\begin{axis}[
			ymax=1,ymin=0,bar width=20,enlarge x limits=0.25,
			symbolic x coords={M,W,T,S,P},
			xtick=data,xtick pos=left,ymajorgrids=true,
			ytick pos=left]
			\addplot[ybar,fill=blue!40] coordinates {
(M,0.6524792204069934)
(W,0.6451132129550015)
(S,0.6085411292633992)
(T,0.6337919174548581)
(P,0.35786758383490974)
			};
			\end{axis}
			\end{tikzpicture}
		}
		\caption{Recall}
		\label{fig:eval_baselines_recall_ru}
	\end{subfigure}
	\begin{subfigure}{.23\linewidth}
		\centering
		\huge
		\resizebox{\linewidth}{!}{
			\begin{tikzpicture}[scale=1]
			\begin{axis}[
			ymax=1,ymin=0,bar width=20,enlarge x limits=0.25,
			symbolic x coords={M,W,T,S,P},
			xtick=data,xtick pos=left,ymajorgrids=true,
			ytick pos=left]
			\addplot[ybar,fill=blue!40] coordinates {
(M,0.8567648612016626)
(W,0.8810518038377024)
(S,0.7556455087912779)
(T,0.8215796585186096)
(P,0.6660294258144644)
			};
			\end{axis}
			\end{tikzpicture}
		}
		\caption{I--Gran}
		\label{fig:eval_baselines_granularity_ru}
	\end{subfigure}
	\begin{subfigure}{.23\linewidth}
		\centering
		\huge
		\resizebox{\linewidth}{!}{
			\begin{tikzpicture}[scale=1]
			\begin{axis}[
			ymax=1,ymin=0,bar width=20,enlarge x limits=0.25,
			symbolic x coords={M,W,T,S,P},
			xtick=data,xtick pos=left,ymajorgrids=true,
			ytick pos=left]
			\addplot[ybar,fill=blue!40] coordinates {
(M,0.6306227015109099)
(W,0.5949043643747116)
(S,0.5534205513442884)
(T,0.5504163526528771)
(P,0.30895534824927473)
			};
			\end{axis}
			\end{tikzpicture}
		}
		\caption{Plagdet}
		\label{fig:eval_baselines_plagdet_ru}
	\end{subfigure}
	
	\caption{Evaluation metric scores for two language pairs, different metrics, alignment methods and baselines. M: \textit{MultiWiki}, W: \textit{WikiParagraphs}, T: \textit{TextTiling}, S: \textit{SA Baseline}, P: \textit{PD Baseline}.}\label{fig:eval_baselines}
\end{figure}

%% file: parameter_diagram.tex
\begin{figure*}[th!]

	\begin{subfigure}{.48\linewidth}
		\centering
		\LARGE
		\resizebox{0.8\linewidth}{!}{ 

			\begin{tikzpicture}
			\begin{axis}[legend pos=south west,
			ymax=1,ymin=0,xmax=0.7,xmin=0.0,
			    ]
			\addplot[color=black,ultra  thick] coordinates {
(0.0,0.5878579413976341)
(0.005,0.6015357773726296)
(0.01,0.6080979236530447)
(0.015,0.5904142664399366)
(0.02,0.6020206444965966)
(0.025,0.6126696406679647)
(0.030000000000000002,0.6107914180755016)
(0.035,0.6117772437572587)
(0.04,0.6181003419091472)
(0.045,0.6131388069276376)
(0.049999999999999996,0.6266960772616894)
(0.05499999999999999,0.6335473552615761)
(0.05999999999999999,0.6419622428659828)
(0.06499999999999999,0.6454791893877813)
(0.06999999999999999,0.6511571652174137)
(0.075,0.6649782530119829)
(0.08,0.6591744361082189)
(0.085,0.6714770680231186)
(0.09000000000000001,0.6750632186651734)
(0.09500000000000001,0.6813558320301853)
(0.10000000000000002,0.6777119368953644)
(0.10500000000000002,0.6901153907373734)
(0.11000000000000003,0.7122780123499018)
(0.11500000000000003,0.7145532768393918)
(0.12000000000000004,0.7155971928389943)
(0.12500000000000003,0.7382701753675502)
(0.13000000000000003,0.7581446812321001)
(0.13500000000000004,0.7687426005687003)
(0.14000000000000004,0.7609504744344386)
(0.14500000000000005,0.7655098200522479)
(0.15000000000000005,0.7788555664690368)
(0.15500000000000005,0.8012213755926758)
(0.16000000000000006,0.8176668738651819)
(0.16500000000000006,0.8138133902282709)
(0.17000000000000007,0.8117483377312217)
(0.17500000000000007,0.8376161385978664)
(0.18000000000000008,0.8425790907561561)
(0.18500000000000008,0.8628138820831485)
(0.19000000000000009,0.8688632974638639)
(0.1950000000000001,0.8688632974638639)
(0.2000000000000001,0.879064420289934)
(0.2050000000000001,0.879064420289934)
(0.2100000000000001,0.8850953156178919)
(0.2150000000000001,0.8983780825297593)
(0.2200000000000001,0.8983780825297593)
(0.22500000000000012,0.8956720022211553)
(0.23000000000000012,0.9027778597300175)
(0.23500000000000013,0.9099692875889495)
(0.24000000000000013,0.9134619915465609)
(0.24500000000000013,0.9116984062449917)
(0.2500000000000001,0.9116984062449917)
(0.2550000000000001,0.9116984062449917)
(0.2600000000000001,0.9172138527195316)
(0.2650000000000001,0.9124025665333982)
(0.27000000000000013,0.9163437363614249)
(0.27500000000000013,0.9205717451909915)
(0.28000000000000014,0.9062625438392232)
(0.28500000000000014,0.9101490745677183)
(0.29000000000000015,0.9049906148834231)
(0.29500000000000015,0.9049906148834231)
(0.30000000000000016,0.9043385527466242)
(0.30500000000000016,0.9036681447542018)
(0.31000000000000016,0.9036681447542018)
(0.31500000000000017,0.9073362902089488)
(0.3200000000000002,0.9127084022395856)
(0.3250000000000002,0.9079188115644247)
(0.3300000000000002,0.9079188115644247)
(0.3350000000000002,0.9048833992170504)
(0.3400000000000002,0.9301644830463216)
(0.3450000000000002,0.9301644830463216)
(0.3500000000000002,0.9285173043247368)
(0.3550000000000002,0.9297787957150515)
(0.3600000000000002,0.923936735555957)
(0.3650000000000002,0.9112010720501852)
(0.3700000000000002,0.9112010720501852)
(0.3750000000000002,0.9159442200299119)
(0.3800000000000002,0.9159442200299119)
(0.38500000000000023,0.9159442200299119)
(0.39000000000000024,0.9159442200299119)
(0.39500000000000024,0.9117761913035136)
(0.40000000000000024,0.9117761913035136)
(0.40500000000000025,0.9146370622397686)
(0.41000000000000025,0.8998689538713274)
(0.41500000000000026,0.8998689538713274)
(0.42000000000000026,0.8998689538713274)
(0.42500000000000027,0.8998689538713274)
(0.43000000000000027,0.8931603838637364)
(0.4350000000000003,0.8931603838637364)
(0.4400000000000003,0.8931603838637364)
(0.4450000000000003,0.9094359465023415)
(0.4500000000000003,0.9094359465023415)
(0.4550000000000003,0.9094359465023415)
(0.4600000000000003,0.9094359465023415)
(0.4650000000000003,0.9094359465023415)
(0.4700000000000003,0.90778836355779)
(0.4750000000000003,0.90778836355779)
(0.4800000000000003,0.9047930537973949)
(0.4850000000000003,0.9047930537973949)
(0.4900000000000003,0.9047930537973949)
(0.49500000000000033,0.9109692560453088)
(0.5000000000000003,0.9109692560453088)
(0.5050000000000003,0.9109692560453088)
(0.5100000000000003,0.9070168212758966)
(0.5150000000000003,0.9070168212758966)
(0.5200000000000004,0.9070168212758966)
(0.5250000000000004,0.9070168212758966)
(0.5300000000000004,0.9088858685833158)
(0.5350000000000004,0.9088858685833158)
(0.5400000000000004,0.9119347078275443)
(0.5450000000000004,0.9119347078275443)
(0.5500000000000004,0.9250819083091818)
(0.5550000000000004,0.9250819083091818)
(0.5600000000000004,0.9250819083091818)
(0.5650000000000004,0.9711936883613592)
(0.5700000000000004,0.9685595192551658)
(0.5750000000000004,0.9685595192551658)
(0.5800000000000004,0.964322434305018)
(0.5850000000000004,0.9625174816540513)
(0.5900000000000004,0.9625174816540513)
(0.5950000000000004,0.9625174816540513)
(0.6000000000000004,0.9625174816540513)
(0.6050000000000004,0.9582840094143155)
(0.6100000000000004,0.95260036158401)
(0.6150000000000004,0.9690740049660937)
(0.6200000000000004,0.9708548910598938)
(0.6250000000000004,0.9708548910598938)
(0.6300000000000004,0.9708548910598938)
(0.6350000000000005,0.9708548910598938)
(0.6400000000000005,0.9708548910598938)
(0.6450000000000005,0.9708548910598938)
(0.6500000000000005,0.9708548910598938)
(0.6550000000000005,0.9708548910598938)
(0.6600000000000005,0.9708548910598938)
(0.6650000000000005,0.9708548910598938)
(0.6700000000000005,0.9708548910598938)
(0.6750000000000005,0.965451309620495)
(0.6800000000000005,0.965451309620495)
(0.6850000000000005,0.965451309620495)
(0.6900000000000005,0.9592257040273564)
(0.6950000000000005,0.9452366503558909)
(0.7000000000000005,0.9452366503558909)
(0.7050000000000005,0.9452366503558909)
(0.7100000000000005,0.9452366503558909)
(0.7150000000000005,0.9452366503558909)
(0.7200000000000005,0.9452366503558909)
(0.7250000000000005,0.9452366503558909)
(0.7300000000000005,0.9452366503558909)
(0.7350000000000005,0.9452366503558909)
(0.7400000000000005,0.9452366503558909)
(0.7450000000000006,0.9452366503558909)
(0.7500000000000006,0.9452366503558909)
(0.7550000000000006,0.9452366503558909)
(0.7600000000000006,0.9452366503558909)
(0.7650000000000006,0.9452366503558909)
(0.7700000000000006,1.0)
(0.7750000000000006,1.0)
(0.7800000000000006,1.0)
(0.7850000000000006,1.0)
(0.7900000000000006,1.0)
(0.7950000000000006,1.0)
(0.8000000000000006,1.0)
(0.8050000000000006,1.0)
(0.8100000000000006,1.0)
(0.8150000000000006,1.0)
(0.8200000000000006,1.0)
(0.8250000000000006,1.0)
(0.8300000000000006,1.0)
(0.8350000000000006,1.0)
(0.8400000000000006,1.0)
(0.8450000000000006,1.0)
(0.8500000000000006,1.0)
(0.8550000000000006,1.0)
(0.8600000000000007,1.0)
(0.8650000000000007,1.0)
(0.8700000000000007,1.0)
(0.8750000000000007,1.0)
(0.8800000000000007,1.0)
(0.8850000000000007,1.0)
(0.8900000000000007,1.0)
(0.8950000000000007,1.0)
(0.9000000000000007,1.0)
(0.9050000000000007,1.0)
(0.9100000000000007,1.0)
(0.9150000000000007,1.0)
(0.9200000000000007,1.0)
(0.9250000000000007,1.0)
(0.9300000000000007,1.0)
(0.9350000000000007,1.0)
(0.9400000000000007,1.0)
(0.9450000000000007,1.0)
(0.9500000000000007,1.0)
(0.9550000000000007,1.0)
(0.9600000000000007,1.0)
(0.9650000000000007,1.0)
(0.9700000000000008,1.0)
(0.9750000000000008,1.0)
(0.9800000000000008,1.0)
(0.9850000000000008,1.0)
(0.9900000000000008,1.0)
(0.9950000000000008,1.0)

			};
			  \addlegendentry{Precision}
			\addplot[color=green,ultra  thick] coordinates {
(0.0,0.6612044977824502)
(0.005,0.6682816490872797)
(0.01,0.6879469668752983)
(0.015,0.6675196032318623)
(0.02,0.6769754027225348)
(0.025,0.6912545100042297)
(0.030000000000000002,0.6944852480743278)
(0.035,0.6750226524418593)
(0.04,0.6817005375051395)
(0.045,0.6662752529791675)
(0.049999999999999996,0.6824957640960353)
(0.05499999999999999,0.6832738342620346)
(0.05999999999999999,0.6830525774886677)
(0.06499999999999999,0.6707953495114339)
(0.06999999999999999,0.6824577628423707)
(0.075,0.682549946254433)
(0.08,0.6781178869768832)
(0.085,0.6794062077523001)
(0.09000000000000001,0.6892265218092676)
(0.09500000000000001,0.7051858711798183)
(0.10000000000000002,0.7066420568042034)
(0.10500000000000002,0.7020096665839617)
(0.11000000000000003,0.715228589786902)
(0.11500000000000003,0.715228589786902)
(0.12000000000000004,0.715228589786902)
(0.12500000000000003,0.7260188859586885)
(0.13000000000000003,0.7348962330939568)
(0.13500000000000004,0.7305894638543848)
(0.14000000000000004,0.7264156132642947)
(0.14500000000000005,0.7224592177215037)
(0.15000000000000005,0.7095709945630199)
(0.15500000000000005,0.7015535926456757)
(0.16000000000000006,0.7113308404750285)
(0.16500000000000006,0.7067771610575404)
(0.17000000000000007,0.7028207655147495)
(0.17500000000000007,0.7158149737335268)
(0.18000000000000008,0.7062800970914508)
(0.18500000000000008,0.7069248348307414)
(0.19000000000000009,0.7069248348307414)
(0.1950000000000001,0.7069248348307414)
(0.2000000000000001,0.7124417741813631)
(0.2050000000000001,0.7124417741813631)
(0.2100000000000001,0.7124417741813631)
(0.2150000000000001,0.7035642451503819)
(0.2200000000000001,0.7035642451503819)
(0.22500000000000012,0.7035642451503819)
(0.23000000000000012,0.6845445042375168)
(0.23500000000000013,0.6731177303968923)
(0.24000000000000013,0.6504230169066747)
(0.24500000000000013,0.6370523959757843)
(0.2500000000000001,0.6370523959757843)
(0.2550000000000001,0.6370523959757843)
(0.2600000000000001,0.6324205591293135)
(0.2650000000000001,0.6292629390885656)
(0.27000000000000013,0.609689591654794)
(0.27500000000000013,0.6055213387925479)
(0.28000000000000014,0.5934406190742013)
(0.28500000000000014,0.5887126779731107)
(0.29000000000000015,0.5632011863606993)
(0.29500000000000015,0.5632011863606993)
(0.30000000000000016,0.556693617193556)
(0.30500000000000016,0.5461903208008094)
(0.31000000000000016,0.5461903208008094)
(0.31500000000000017,0.5415897151752631)
(0.3200000000000002,0.5307303411828951)
(0.3250000000000002,0.5217720993411432)
(0.3300000000000002,0.5144463042477594)
(0.3350000000000002,0.5041266068342424)
(0.3400000000000002,0.46635990304016983)
(0.3450000000000002,0.4608939073203387)
(0.3500000000000002,0.44693516886867757)
(0.3550000000000002,0.44464781953653115)
(0.3600000000000002,0.4183274924550791)
(0.3650000000000002,0.3534275357855852)
(0.3700000000000002,0.33937906476661944)
(0.3750000000000002,0.335919241356679)
(0.3800000000000002,0.335919241356679)
(0.38500000000000023,0.335919241356679)
(0.39000000000000024,0.32732881550667325)
(0.39500000000000024,0.3083899159421867)
(0.40000000000000024,0.3083899159421867)
(0.40500000000000025,0.2974041473190507)
(0.41000000000000025,0.27114826552794774)
(0.41500000000000026,0.265329469550468)
(0.42000000000000026,0.26290265817278363)
(0.42500000000000027,0.2588269515737142)
(0.43000000000000027,0.2574605792956632)
(0.4350000000000003,0.2574605792956632)
(0.4400000000000003,0.25131286543114517)
(0.4450000000000003,0.2432634331860727)
(0.4500000000000003,0.2432634331860727)
(0.4550000000000003,0.2432634331860727)
(0.4600000000000003,0.239158266276158)
(0.4650000000000003,0.239158266276158)
(0.4700000000000003,0.23226523133953256)
(0.4750000000000003,0.22503610264345633)
(0.4800000000000003,0.20496581873461076)
(0.4850000000000003,0.1999443697306201)
(0.4900000000000003,0.1999443697306201)
(0.49500000000000033,0.19081999543181963)
(0.5000000000000003,0.19081999543181963)
(0.5050000000000003,0.19081999543181963)
(0.5100000000000003,0.16501340782972157)
(0.5150000000000003,0.16501340782972157)
(0.5200000000000004,0.16501340782972157)
(0.5250000000000004,0.16501340782972157)
(0.5300000000000004,0.16331930672232373)
(0.5350000000000004,0.15596845891180233)
(0.5400000000000004,0.14812474225125163)
(0.5450000000000004,0.14812474225125163)
(0.5500000000000004,0.1436559430200683)
(0.5550000000000004,0.1436559430200683)
(0.5600000000000004,0.1436559430200683)
(0.5650000000000004,0.13295798522599836)
(0.5700000000000004,0.11658029674991341)
(0.5750000000000004,0.11658029674991341)
(0.5800000000000004,0.10959656398517445)
(0.5850000000000004,0.10712504816908178)
(0.5900000000000004,0.10712504816908178)
(0.5950000000000004,0.09855976474202773)
(0.6000000000000004,0.09855976474202773)
(0.6050000000000004,0.09363940756687074)
(0.6100000000000004,0.08846963136440165)
(0.6150000000000004,0.0870134457400166)
(0.6200000000000004,0.07534352021081873)
(0.6250000000000004,0.07534352021081873)
(0.6300000000000004,0.07534352021081873)
(0.6350000000000005,0.07534352021081873)
(0.6400000000000005,0.07212289202734976)
(0.6450000000000005,0.07212289202734976)
(0.6500000000000005,0.07212289202734976)
(0.6550000000000005,0.061797559695569326)
(0.6600000000000005,0.056105041469847225)
(0.6650000000000005,0.056105041469847225)
(0.6700000000000005,0.056105041469847225)
(0.6750000000000005,0.05109398646657675)
(0.6800000000000005,0.05109398646657675)
(0.6850000000000005,0.05109398646657675)
(0.6900000000000005,0.04696723528741282)
(0.6950000000000005,0.041117225374092535)
(0.7000000000000005,0.03421474284900018)
(0.7050000000000005,0.03421474284900018)
(0.7100000000000005,0.03421474284900018)
(0.7150000000000005,0.03421474284900018)
(0.7200000000000005,0.02586832614817916)
(0.7250000000000005,0.02586832614817916)
(0.7300000000000005,0.02586832614817916)
(0.7350000000000005,0.02586832614817916)
(0.7400000000000005,0.021043120389165924)
(0.7450000000000006,0.021043120389165924)
(0.7500000000000006,0.021043120389165924)
(0.7550000000000006,0.021043120389165924)
(0.7600000000000006,0.021043120389165924)
(0.7650000000000006,0.021043120389165924)
(0.7700000000000006,0.013737872050853632)
(0.7750000000000006,0.013737872050853632)
(0.7800000000000006,0.007751815395363105)
(0.7850000000000006,0.007751815395363105)
(0.7900000000000006,0.007751815395363105)
(0.7950000000000006,0.007751815395363105)
(0.8000000000000006,0.007751815395363105)
(0.8050000000000006,0.007751815395363105)
(0.8100000000000006,0.007751815395363105)
(0.8150000000000006,0.007751815395363105)
(0.8200000000000006,0.007751815395363105)
(0.8250000000000006,0.007751815395363105)
(0.8300000000000006,0.007751815395363105)
(0.8350000000000006,0.007751815395363105)
(0.8400000000000006,0.007751815395363105)
(0.8450000000000006,0.007751815395363105)
(0.8500000000000006,0.001746396972726138)
(0.8550000000000006,0.001746396972726138)
(0.8600000000000007,0.001746396972726138)
(0.8650000000000007,0.001746396972726138)
(0.8700000000000007,0.001746396972726138)
(0.8750000000000007,0.001746396972726138)
(0.8800000000000007,0.001746396972726138)
(0.8850000000000007,0.001746396972726138)
(0.8900000000000007,0.0)
(0.8950000000000007,0.0)
(0.9000000000000007,0.0)
(0.9050000000000007,0.0)
(0.9100000000000007,0.0)
(0.9150000000000007,0.0)
(0.9200000000000007,0.0)
(0.9250000000000007,0.0)
(0.9300000000000007,0.0)
(0.9350000000000007,0.0)
(0.9400000000000007,0.0)
(0.9450000000000007,0.0)
(0.9500000000000007,0.0)
(0.9550000000000007,0.0)
(0.9600000000000007,0.0)
(0.9650000000000007,0.0)
(0.9700000000000008,0.0)
(0.9750000000000008,0.0)
(0.9800000000000008,0.0)
(0.9850000000000008,0.0)
(0.9900000000000008,0.0)
(0.9950000000000008,0.0)
			};
			  \addlegendentry{Recall}
			  			\addplot[dotted,color=orange,ultra  thick] coordinates {
(0.0,0.878707325140717)
(0.005,0.8658351948976949)
(0.01,0.8694648604706551)
(0.015,0.8681152845039434)
(0.02,0.8752548768436509)
(0.025,0.8593581129135766)
(0.030000000000000002,0.8593581129135766)
(0.035,0.8725593057666888)
(0.04,0.8630322763145611)
(0.045,0.8725593057666888)
(0.049999999999999996,0.8691605559175923)
(0.05499999999999999,0.8662794366309008)
(0.05999999999999999,0.8662794366309008)
(0.06499999999999999,0.8614683778632454)
(0.06999999999999999,0.8662230127151485)
(0.075,0.8653801213833014)
(0.08,0.8650410331926401)
(0.085,0.8630447242368139)
(0.09000000000000001,0.8621108448578574)
(0.09500000000000001,0.8586392907726857)
(0.10000000000000002,0.8599083046680501)
(0.10500000000000002,0.8567363936826045)
(0.11000000000000003,0.8504130420594523)
(0.11500000000000003,0.8504130420594523)
(0.12000000000000004,0.8504130420594523)
(0.12500000000000003,0.8494295389017847)
(0.13000000000000003,0.848576766579965)
(0.13500000000000004,0.8411901307727458)
(0.14000000000000004,0.8427081746536534)
(0.14500000000000005,0.8439570667642078)
(0.15000000000000005,0.8411566494680222)
(0.15500000000000005,0.8352788106935677)
(0.16000000000000006,0.8328365327884836)
(0.16500000000000006,0.8342085866605375)
(0.17000000000000007,0.8354019724550672)
(0.17500000000000007,0.8342085866605375)
(0.18000000000000008,0.8372888493750568)
(0.18500000000000008,0.8374129430105892)
(0.19000000000000009,0.8374129430105892)
(0.1950000000000001,0.8374129430105892)
(0.2000000000000001,0.8342892896220352)
(0.2050000000000001,0.8342892896220352)
(0.2100000000000001,0.8342892896220352)
(0.2150000000000001,0.8318801026598681)
(0.2200000000000001,0.8318801026598681)
(0.22500000000000012,0.8318801026598681)
(0.23000000000000012,0.832628522286599)
(0.23500000000000013,0.832628522286599)
(0.24000000000000013,0.8338204483784348)
(0.24500000000000013,0.8337760101867961)
(0.2500000000000001,0.8337760101867961)
(0.2550000000000001,0.8337760101867961)
(0.2600000000000001,0.8324680631669287)
(0.2650000000000001,0.8324680631669287)
(0.27000000000000013,0.8447699037700545)
(0.27500000000000013,0.8447699037700545)
(0.28000000000000014,0.8447699037700545)
(0.28500000000000014,0.842605301089582)
(0.29000000000000015,0.8397931693287746)
(0.29500000000000015,0.8397931693287746)
(0.30000000000000016,0.8411310504416315)
(0.30500000000000016,0.8443294612939407)
(0.31000000000000016,0.8443294612939407)
(0.31500000000000017,0.8443294612939407)
(0.3200000000000002,0.8414714105882078)
(0.3250000000000002,0.8521475985160514)
(0.3300000000000002,0.85200907197024)
(0.3350000000000002,0.8572408600497102)
(0.3400000000000002,0.8579676923561567)
(0.3450000000000002,0.865281893974994)
(0.3500000000000002,0.8735270704320911)
(0.3550000000000002,0.8752290420094035)
(0.3600000000000002,0.8822792806780196)
(0.3650000000000002,0.8939547154424422)
(0.3700000000000002,0.8939547154424422)
(0.3750000000000002,0.8968127661481751)
(0.3800000000000002,0.8968127661481751)
(0.38500000000000023,0.8968127661481751)
(0.39000000000000024,0.901802230326381)
(0.39500000000000024,0.9030524440158882)
(0.40000000000000024,0.9030524440158882)
(0.40500000000000025,0.8998891050303419)
(0.41000000000000025,0.909355565211482)
(0.41500000000000026,0.9042833195763225)
(0.42000000000000026,0.9042833195763225)
(0.42500000000000027,0.9042833195763225)
(0.43000000000000027,0.9042833195763225)
(0.4350000000000003,0.9042833195763225)
(0.4400000000000003,0.8344827476317107)
(0.4450000000000003,0.8373390050608158)
(0.4500000000000003,0.8373390050608158)
(0.4550000000000003,0.8373390050608158)
(0.4600000000000003,0.8408298740152609)
(0.4650000000000003,0.8408298740152609)
(0.4700000000000003,0.8421348126518179)
(0.4750000000000003,0.8461553575043164)
(0.4800000000000003,0.8480210532130348)
(0.4850000000000003,0.8480210532130348)
(0.4900000000000003,0.8480210532130348)
(0.49500000000000033,0.8120540989017097)
(0.5000000000000003,0.8120540989017097)
(0.5050000000000003,0.8120540989017097)
(0.5100000000000003,0.8259458711908069)
(0.5150000000000003,0.8259458711908069)
(0.5200000000000004,0.8259458711908069)
(0.5250000000000004,0.8259458711908069)
(0.5300000000000004,0.8259458711908069)
(0.5350000000000004,0.831090362461126)
(0.5400000000000004,0.8392635476493092)
(0.5450000000000004,0.8392635476493092)
(0.5500000000000004,0.8182280268786891)
(0.5550000000000004,0.8182280268786891)
(0.5600000000000004,0.8182280268786891)
(0.5650000000000004,0.7733971296057389)
(0.5700000000000004,0.7770194618753247)
(0.5750000000000004,0.7770194618753247)
(0.5800000000000004,0.781983545642486)
(0.5850000000000004,0.7931736450941748)
(0.5900000000000004,0.7931736450941748)
(0.5950000000000004,0.7931736450941748)
(0.6000000000000004,0.7931736450941748)
(0.6050000000000004,0.7919933734292554)
(0.6100000000000004,0.7902885365799273)
(0.6150000000000004,0.7902885365799273)
(0.6200000000000004,0.7419122237830845)
(0.6250000000000004,0.7419122237830845)
(0.6300000000000004,0.7419122237830845)
(0.6350000000000005,0.7419122237830845)
(0.6400000000000005,0.7419122237830845)
(0.6450000000000005,0.7419122237830845)
(0.6500000000000005,0.7419122237830845)
(0.6550000000000005,0.7385888409481124)
(0.6600000000000005,0.6819111367538518)
(0.6650000000000005,0.6819111367538518)
(0.6700000000000005,0.6819111367538518)
(0.6750000000000005,0.6912877394251565)
(0.6800000000000005,0.6912877394251565)
(0.6850000000000005,0.6912877394251565)
(0.6900000000000005,0.6912877394251565)
(0.6950000000000005,0.6912877394251565)
(0.7000000000000005,0.671329882712333)
(0.7050000000000005,0.671329882712333)
(0.7100000000000005,0.671329882712333)
(0.7150000000000005,0.671329882712333)
(0.7200000000000005,0.671329882712333)
(0.7250000000000005,0.671329882712333)
(0.7300000000000005,0.671329882712333)
(0.7350000000000005,0.671329882712333)
(0.7400000000000005,0.6586154034829531)
(0.7450000000000006,0.6586154034829531)
(0.7500000000000006,0.6586154034829531)
(0.7550000000000006,0.6586154034829531)
(0.7600000000000006,0.6586154034829531)
(0.7650000000000006,0.6586154034829531)
(0.7700000000000006,0.6586154034829531)
(0.7750000000000006,0.6586154034829531)
(0.7800000000000006,0.4383991921740266)
(0.7850000000000006,0.4383991921740266)
(0.7900000000000006,0.4383991921740266)
(0.7950000000000006,0.4383991921740266)
(0.8000000000000006,0.4383991921740266)
(0.8050000000000006,0.4383991921740266)
(0.8100000000000006,0.4383991921740266)
(0.8150000000000006,0.4383991921740266)
(0.8200000000000006,0.4383991921740266)
(0.8250000000000006,0.4383991921740266)
(0.8300000000000006,0.4383991921740266)
(0.8350000000000006,0.4383991921740266)
(0.8400000000000006,0.4383991921740266)
(0.8450000000000006,0.4383991921740266)
(0.8500000000000006,0.42097683114653317)
(0.8550000000000006,0.42097683114653317)
(0.8600000000000007,0.42097683114653317)
(0.8650000000000007,0.42097683114653317)
(0.8700000000000007,0.42097683114653317)
(0.8750000000000007,0.42097683114653317)
(0.8800000000000007,0.42097683114653317)
(0.8850000000000007,0.42097683114653317)
(0.8900000000000007,0.39130444503374967)
(0.8950000000000007,0.39130444503374967)
(0.9000000000000007,0.39130444503374967)
(0.9050000000000007,0.39130444503374967)
(0.9100000000000007,0.39130444503374967)
(0.9150000000000007,0.39130444503374967)
(0.9200000000000007,0.39130444503374967)
(0.9250000000000007,0.39130444503374967)
(0.9300000000000007,0.39130444503374967)
(0.9350000000000007,0.39130444503374967)
(0.9400000000000007,0.39130444503374967)
(0.9450000000000007,0.39130444503374967)
(0.9500000000000007,0.39130444503374967)
(0.9550000000000007,0.39130444503374967)
(0.9600000000000007,0.39130444503374967)
(0.9650000000000007,0.39130444503374967)
(0.9700000000000008,0.39130444503374967)
(0.9750000000000008,0.39130444503374967)
(0.9800000000000008,0.39130444503374967)
(0.9850000000000008,0.39130444503374967)
(0.9900000000000008,0.39130444503374967)
(0.9950000000000008,0.39130444503374967)
			};
			  \addlegendentry{I--Gran}	
			  		\addplot[dashed,color=blue,ultra  thick] coordinates {
(0.0,0.547533659915495)
(0.005,0.550911048387207)
(0.01,0.5634264276981745)
(0.015,0.545562109421405)
(0.02,0.5577874319202205)
(0.025,0.5668857766371255)
(0.030000000000000002,0.5662794669978652)
(0.035,0.5649696169888905)
(0.04,0.567463349790445)
(0.045,0.5631203696054848)
(0.049999999999999996,0.5749591824863942)
(0.05499999999999999,0.5781108601415864)
(0.05999999999999999,0.5822319006721183)
(0.06499999999999999,0.5780022143538703)
(0.06999999999999999,0.585296189290336)
(0.075,0.5925408644416448)
(0.08,0.5869698837768066)
(0.085,0.59222030746041)
(0.09000000000000001,0.5970288876020396)
(0.09500000000000001,0.6070402633249435)
(0.10000000000000002,0.6063743061925538)
(0.10500000000000002,0.606307252673644)
(0.11000000000000003,0.6161192340062239)
(0.11500000000000003,0.6172576613181033)
(0.12000000000000004,0.6177545255907451)
(0.12500000000000003,0.6312169077303131)
(0.13000000000000003,0.6421047053193673)
(0.13500000000000004,0.6397774946547112)
(0.14000000000000004,0.6350810775157464)
(0.14500000000000005,0.6365143383405056)
(0.15000000000000005,0.6247886554746784)
(0.15500000000000005,0.6226897262153689)
(0.16000000000000006,0.6319113539450801)
(0.16500000000000006,0.6284259119159552)
(0.17000000000000007,0.6269213317434349)
(0.17500000000000007,0.6439455416497749)
(0.18000000000000008,0.6429781737546049)
(0.18500000000000008,0.645819101169328)
(0.19000000000000009,0.6479924891700474)
(0.1950000000000001,0.6479924891700474)
(0.2000000000000001,0.6533091871774614)
(0.2050000000000001,0.6533091871774614)
(0.2100000000000001,0.6551838870315086)
(0.2150000000000001,0.6503462949607554)
(0.2200000000000001,0.6503462949607554)
(0.22500000000000012,0.6493789761534094)
(0.23000000000000012,0.6436132847094903)
(0.23500000000000013,0.6400400974827717)
(0.24000000000000013,0.6250951129214757)
(0.24500000000000013,0.6182522219518084)
(0.2500000000000001,0.6182522219518084)
(0.2550000000000001,0.6182522219518084)
(0.2600000000000001,0.6174697279847006)
(0.2650000000000001,0.6138307057303355)
(0.27000000000000013,0.6080601316196202)
(0.27500000000000013,0.6062286165625812)
(0.28000000000000014,0.5923187285782364)
(0.28500000000000014,0.5884363469591212)
(0.29000000000000015,0.5650496792742505)
(0.29500000000000015,0.5650496792742505)
(0.30000000000000016,0.5625564451361992)
(0.30500000000000016,0.5556488714002036)
(0.31000000000000016,0.5556488714002036)
(0.31500000000000017,0.5536842147560856)
(0.3200000000000002,0.546881496691777)
(0.3250000000000002,0.5428357439652267)
(0.3300000000000002,0.537974136879511)
(0.3350000000000002,0.5293938519072541)
(0.3400000000000002,0.5063924966279328)
(0.3450000000000002,0.5056661000445276)
(0.3500000000000002,0.49569712549383976)
(0.3550000000000002,0.4957141978712709)
(0.3600000000000002,0.47245823261929687)
(0.3650000000000002,0.42812632477173174)
(0.3700000000000002,0.40968257503091754)
(0.3750000000000002,0.4107267838906019)
(0.3800000000000002,0.4107267838906019)
(0.38500000000000023,0.4107267838906019)
(0.39000000000000024,0.39994765631325574)
(0.39500000000000024,0.38556543867372195)
(0.40000000000000024,0.38556543867372195)
(0.40500000000000025,0.3734948961446665)
(0.41000000000000025,0.34837686284771624)
(0.41500000000000026,0.3400359457023427)
(0.42000000000000026,0.3377982072012026)
(0.42500000000000027,0.3308674080028916)
(0.43000000000000027,0.32858231085299594)
(0.4350000000000003,0.32858231085299594)
(0.4400000000000003,0.31702256255789285)
(0.4450000000000003,0.31219696498381533)
(0.4500000000000003,0.31219696498381533)
(0.4550000000000003,0.31219696498381533)
(0.4600000000000003,0.30939771223088747)
(0.4650000000000003,0.30939771223088747)
(0.4700000000000003,0.3034849354381466)
(0.4750000000000003,0.30077534725675026)
(0.4800000000000003,0.282829048859232)
(0.4850000000000003,0.2746233910761877)
(0.4900000000000003,0.2746233910761877)
(0.49500000000000033,0.2591482912975116)
(0.5000000000000003,0.2591482912975116)
(0.5050000000000003,0.2591482912975116)
(0.5100000000000003,0.23949559914650426)
(0.5150000000000003,0.23949559914650426)
(0.5200000000000004,0.23949559914650426)
(0.5250000000000004,0.23949559914650426)
(0.5300000000000004,0.2378447147924263)
(0.5350000000000004,0.23184480483427491)
(0.5400000000000004,0.22525479749922275)
(0.5450000000000004,0.22525479749922275)
(0.5500000000000004,0.21773812535504541)
(0.5550000000000004,0.21773812535504541)
(0.5600000000000004,0.21773812535504541)
(0.5650000000000004,0.20136625873982372)
(0.5700000000000004,0.18230365408552845)
(0.5750000000000004,0.18230365408552845)
(0.5800000000000004,0.1742147729394893)
(0.5850000000000004,0.17282766375481626)
(0.5900000000000004,0.17282766375481626)
(0.5950000000000004,0.16324274375171516)
(0.6000000000000004,0.16324274375171516)
(0.6050000000000004,0.1558438045841084)
(0.6100000000000004,0.14778712649803968)
(0.6150000000000004,0.1459709224876458)
(0.6200000000000004,0.12544041773121598)
(0.6250000000000004,0.12544041773121598)
(0.6300000000000004,0.12544041773121598)
(0.6350000000000005,0.12544041773121598)
(0.6400000000000005,0.12113993361787412)
(0.6450000000000005,0.12113993361787412)
(0.6500000000000005,0.12113993361787412)
(0.6550000000000005,0.10746657169041375)
(0.6600000000000005,0.09686838097678682)
(0.6650000000000005,0.09686838097678682)
(0.6700000000000005,0.09686838097678682)
(0.6750000000000005,0.09012081638667584)
(0.6800000000000005,0.09012081638667584)
(0.6850000000000005,0.09012081638667584)
(0.6900000000000005,0.08318336084473935)
(0.6950000000000005,0.07298468066496998)
(0.7000000000000005,0.06060820549561973)
(0.7050000000000005,0.06060820549561973)
(0.7100000000000005,0.06060820549561973)
(0.7150000000000005,0.06060820549561973)
(0.7200000000000005,0.04707450437495357)
(0.7250000000000005,0.04707450437495357)
(0.7300000000000005,0.04707450437495357)
(0.7350000000000005,0.04707450437495357)
(0.7400000000000005,0.03896678735402088)
(0.7450000000000006,0.03896678735402088)
(0.7500000000000006,0.03896678735402088)
(0.7550000000000006,0.03896678735402088)
(0.7600000000000006,0.03896678735402088)
(0.7650000000000006,0.03896678735402088)
(0.7700000000000006,0.02575535576994865)
(0.7750000000000006,0.02575535576994865)
(0.7800000000000006,0.014060884626851868)
(0.7850000000000006,0.014060884626851868)
(0.7900000000000006,0.014060884626851868)
(0.7950000000000006,0.014060884626851868)
(0.8000000000000006,0.014060884626851868)
(0.8050000000000006,0.014060884626851868)
(0.8100000000000006,0.014060884626851868)
(0.8150000000000006,0.014060884626851868)
(0.8200000000000006,0.014060884626851868)
(0.8250000000000006,0.014060884626851868)
(0.8300000000000006,0.014060884626851868)
(0.8350000000000006,0.014060884626851868)
(0.8400000000000006,0.014060884626851868)
(0.8450000000000006,0.014060884626851868)
(0.8500000000000006,0.0033795869413127985)
(0.8550000000000006,0.0033795869413127985)
(0.8600000000000007,0.0033795869413127985)
(0.8650000000000007,0.0033795869413127985)
(0.8700000000000007,0.0033795869413127985)
(0.8750000000000007,0.0033795869413127985)
(0.8800000000000007,0.0033795869413127985)
(0.8850000000000007,0.0033795869413127985)
(0.8900000000000007,0.0)
(0.8950000000000007,0.0)
(0.9000000000000007,0.0)
(0.9050000000000007,0.0)
(0.9100000000000007,0.0)
(0.9150000000000007,0.0)
(0.9200000000000007,0.0)
(0.9250000000000007,0.0)
(0.9300000000000007,0.0)
(0.9350000000000007,0.0)
(0.9400000000000007,0.0)
(0.9450000000000007,0.0)
(0.9500000000000007,0.0)
(0.9550000000000007,0.0)
(0.9600000000000007,0.0)
(0.9650000000000007,0.0)
(0.9700000000000008,0.0)
(0.9750000000000008,0.0)
(0.9800000000000008,0.0)
(0.9850000000000008,0.0)
(0.9900000000000008,0.0)
(0.9950000000000008,0.0)
			};			
			  \addlegendentry{Plagdet}
			
			\end{axis}
			\end{tikzpicture}
		}
		\caption{Similarity threshold ($sf=min$)}
		\label{fig:th}
	\end{subfigure} 
	\begin{subfigure}{.48\linewidth}
		\centering
		\LARGE
		\resizebox{0.746\linewidth}{!}{ 
			\begin{tikzpicture}
			\begin{axis}[legend pos=south west,
			ymax=1,ymin=0,bar width=7,enlarge x limits=0.25,
			symbolic x coords={MIN,MID,MAX},
			xtick=data,xtick pos=left,ymajorgrids=true,xbar=5pt,
			ytick pos=left,ybar=5pt,
			]
			\addplot[ybar,color=black,fill=black,ultra thick,area legend ] coordinates {
(MIN,0.8850953156178917)
(MID,0.8351930727587306)
(MAX,0.756258396099135)
			};
\addlegendentry{Precision}			\addplot[ybar,color=green,ultra thick,fill=green,area legend ] coordinates {
(MIN,0.712441774181363)
(MID,0.7519768033128329)
(MAX,0.843505441681842)
			};
\addlegendentry{Recall}			\addplot[ybar,color=orange,ultra thick,fill=orange,area legend ] coordinates {
(MIN,0.834289289622035)
(MID,0.7911873748721843)
(MAX,0.4916235075630771)
			};
\addlegendentry{I--Gran}			\addplot[ybar,color=blue,ultra thick,fill=blue,area legend ] coordinates {
(MIN,0.6551838870315086)
(MID,0.6259266842624942)
(MAX,0.4529106809878041)
			};			
\addlegendentry{Plagdet}			
			
			\end{axis}
			\end{tikzpicture}
		}
		\caption{Structure freedom ($th=0.21$)}
		\label{fig:sf}
	\end{subfigure}

	\caption{Evaluation scores for different parameters used in our algorithm. In both diagrams, these scores are computed
		for varying values of the parameter on the X-axis while the other parameter is
		kept constant.}
	\label{fig:eval_configurations}
\end{figure*}

%% file: 08_related.tex
\section{Related Research}
\label{sec:related}

The problem of the interlingual text passage alignment
in partner articles discussed in this paper has not been addressed by any
existing approach.
In the following we discuss related applications to analyze interlingual
differences in multilingual Wikipedia as well as related methods for 
interlingual text alignment. Finally, we discuss available benchmarks.

\textbf{Analyzing interlingual differences in Wikipedia:}
The problem of identifying information missing in a particular language
edition using other Wikipedia language editions has been considered at different
levels of granularity, including suggestion of the articles missing in a
particular language edition to the Wikipedia editors \cite{Wulczyn:2016}, finding
complementary sentences within a partner article \cite{Duh:2013} and detection
of missing infobox information \cite{Adar:2009}.
Information propagation across languages has been considered by 
\citeN{Hale:2014}, who studied the behavior of the editors
simultaneously working on multiple Wikipedia language editions. Interlingual
information propagation has also been considered in our recent demonstration
paper, where we proposed a graphical user interface to observe changes in
the interlingual article similarity over time \cite{gottschalk2016}.
 All these approaches target discovery of interlingual similarities in
 Wikipedia, while targeting aspects different from MultiWiki.

Further approaches attempt to automatically compare partner articles 
to identify their overall similarity. 
In Barr\'{o}n-Cede\~{n}o et al. \citeyear{Cedeno:2014}, 
the authors compare different metrics to compute an 
overall similarity of the articles in different languages.
Manypedia \cite{Massa2012} provides an automatic translation to English, 
and points out article statistics and
concept similarity metrics computed based on the article interlinking.
The Omnipedia interface  \cite{Bao:2012} visualizes the information summarized
from multiple language editions using topic extraction methods.
However, none of the existing approaches enables the detailed interlingual
comparison of the partner articles at the text passage level as facilitated by
MultiWiki.

\textbf{Interlingual text passage alignment:}
Text passage alignment has been considered in the context of 
machine translation applications, where the goal is to create 
parallel corpora to train translation models.
In this context, existing approaches aim to extract parallel text passages from
a bilingual parallel document corpus 
(e.g. \cite{Rasooli:2011}, \cite{Gupta:2012}). 
Existing approaches in this area typically assume that the paragraphs are 
parallel and contain translated text, such that adjustments 
of the paragraph boundaries are not required. 
Rasooli et al. \citeyear{Rasooli:2011}  use pre-defined paragraph boundaries and
apply similarity measures, similar to the method \textit{WikiParagraphs} used as a
baseline in this paper.
Gupta et al. \citeyear{Gupta:2012} allow for many-to-many paragraph alignments 
(i.e. they merge neighbored paragraphs) based on the assumption 
of a common text flow in both documents. These assumptions are not applicable to 
the partner articles in Wikipedia. In contrast, the MultiWiki method does not
require any parallel corpora and facilitates an alignment of similar text
passages irrespective of the differences in the paragraph structure.

At a higher granularity level, 
several works have also considered an alignment of individual parallel
sentences in the context of 
machine translation \cite{Adafre2006}, \cite{Mohammadi10} 
and identification of complementary sentences in partner
articles \cite{Duh:2013}.
Sentence alignment alone fails to provide an overview of the overlapping article 
parts due to its high granularity.
In addition, in MultiWiki we face the problem of the alignment of partially
overlapping text, that can contain intermediate unrelated sentence parts or entire sentences.
The use of semantic features enables MultiWiki to overcome the limitations
related to the syntactic alignment of parallel sentences and achieve better
recall and granularity of the alignment,
as demonstrated by our experimental evaluation.

\textbf{Interlingual text reuse and plagiarism detection:}
Text reuse occurs for various reasons and can be of different
granularity, including reuse of entire documents as well as extracts thereof,
such as sentences, facts or text passages (also known as local text reuse
\cite{Seo:2008}). In this context, plagiarism detection is a
special form of local text reuse detection.
Interlingual plagiarism detection focuses on identification of reused text
passages across a suspicious document and possible source documents. 
In the first step, plagiarism detection methods try to identify source documents
for a given suspicious input document. Then, they search for 
text fragments that are found both in the suspicious document and the source 
documents \cite{Alzahrani2010}.

There are two different approaches to identify plagiarized text passages: 
(a) Subdividing the text into sections to build a 
tree structure of the document that is used to reduce the number of comparisons 
\cite{Chow:2009}; and (b) 
Bottom-up combination of sentences into text passages \cite{Alzahrani2010}. 
While (a) relies on similar passage partitioning between 
the texts, (b) highly relies on correctly aligned sentences: In \cite{Alzahrani2010}, 
they merge aligned sentences 
that have a distance of at most 10 characters. This is similar to the merging of 
neighbored text passage pairs in our method, but 
lacks the inclusion of the intermediate sentences in an aligned text passage and
there is no similarity re-computation between text passages consisting 
of more than one sentence.

There are two important differences between plagiarism detection and the problem of 
finding similar text passages across Wikipedia article pairs:
First, plagiarism detection is a directed problem: Given a suspicious document and one 
or more source documents, the goal is to search 
for text passages in the suspicious document that are based on the source documents. 
In the Wikipedia text passage alignment, there 
is no direction: Due to the independent evolution of Wikipedia articles in different language editions, both 
partner articles assume the role of a suspicious document and a source document at a time.
Second, a plagiarized text passage must be based on a text passage in another document. 
Therefore, many plagiarism detection 
methods use syntactic similarity measures like n-grams or the 
longest common subsequence \cite{Alzahrani2010} that take the 
order of words or characters into account. In our case, aligned text 
passages share common information 
without necessarily being based on the same source.
Because of these differences, our MultiWiki method allows for larger portions of 
additional information such as unaligned sentences or facts within the aligned text passages.

\textbf{Benchmarks:}
Existing parallel corpora (e.g. \cite{koehn2005epc}, \cite{Steinberger06})
cover parallel sentences from particular domains (e.g. news domain or 
parliamentary proceedings).
Smith et al. \citeyear{Smith:2010} provided a small benchmark with 225 parallel 
sentences extracted from 20
manually chosen Wikipedia articles.
However, existing corpora focus on parallel sentences and
do not include the sentences with partially overlapping information nuggets.
SemEval workshop on semantic evaluation \cite{SemEval2016} includes a 
cross-lingual semantic textual similarity task and provides Spanish-English
bilingual sentence pairs. The sentences within this benchmark come from the domains
different from Wikipedia, are rather short and rarely contain time information, as opposed
to our $Sim-B$ benchmark.
In this work we incrementally build a benchmark $Sim-B$ for the alignment of
sentences with substantial semantic overlap for the German and 
the English Wikipedia.
Furthermore, we collect a user-annotated benchmark $Align-B$ for text passage
extraction and alignment from partner articles.
We make both benchmarks available to facilitate further research in this
area.

%% file: 09_conclusion.tex
\section{Discussion and Future Work}
\label{sec:discussion}

In this article we tackled the problem of
interlingual alignment of semantically similar text passages across partner
articles in Wikipedia.
Partner articles evolve independently in different language editions and can
therefore reflect community-specific points of view on particular topics, or
indicate other differences with respect to the content, structure and quality
of the information they contain.
MultiWiki is the first method that facilitates direct comparison of the
similarities and differences in the interlingual partner article pairs at the
text passage level, providing users with a detailed overview.

\textit{Contributions of the article:} In order to facilitate a comprehensive
overview of the interlingual similarities and differences in an article pair,
we defined text passage alignment as an optimization problem. This problem 
maximizes the semantic similarity across the aligned text passages while 
reducing their granularity. This way, we aim at obtaining possibly
long and semantically similar interlingual text passage pairs.
Then, we designed a method to address this optimization problem and defined a 
semantic similarity measure for the interlingual text passage alignment 
along with a greedy algorithm to perform text passage extraction.
A further contribution of this
article are the user-annotated benchmarks containing aligned text passages from the 
German, Russian and the English Wikipedia language
editions.
Our evaluation results 
on German-English and Russian-English article pairs
demonstrate that our method achieves 
a good balance between precision, recall and granularity of the aligned text
passages as measured against the user annotations. 

\textit{Extensions to other language pairs:} The MultiWiki system is
publicly available\footnote{\url{https://github.com/sgottsch/multiwiki}} 
and its demonstration currently supports four language
pairs: German-English, Dutch-English, Portuguese-English and
Russian-English. The set of the language pairs supported by MultiWiki is
extendible as long as the minimal requirements on the availability of
the language processing tools are satisfied. This includes sentence
splitting, tokenization and machine translation 
for the corresponding language pair.
As machine translation is 
used to obtain the term vector representations of text passages, 
the only requirement on the translation quality is the 
correct translation for the majority of the terms. 
As we observed, entity annotations and time annotations can 
further increase the precision of the interlingual text passage alignment. 
Annotation tools such as DBpedia Spotlight and HeidelTime 
are already available in a number of languages making it possible to further
extend the number of languages supported by MultiWiki in the future.
Another interesting direction
for future research is the reduction of the need for machine translation while
aligning multilingual text by the development of further text similarity
features, e.g. by
utilizing multilingual word embeddings \cite{Vulic:2015} or language-independent word sense 
disambiguation \cite{Pilehvar:2013}, \cite{Moro:2014}.

\textit{Domain adaptation:} In this article we focused on the interlingual text
passage alignment in Wikipedia. Using this corpus, we can utilize its specific features such as
the interlingual links between partner articles, the comparable text styles and
the encyclopedic nature of the articles which enables to extract a relatively
high number of semantic annotations. In our future research we would like to 
consider an adaptation of this approach to other domains, such as multilingual
news and social media, where these features may not be available to the same
extent. Adaptation to these domains may require establishing
interlingual links at the article level, as well as adaptation of the similarity
function and alignment algorithms to better match the 
features and text structure in these domains.

\textit{Exclusiveness in the alignment model:} In our problem
statement we assume the mutual exclusiveness of text passages within the 
article as well as with respect to the alignment. 
The intuition behind this assumption is that such exclusive alignment can 
facilitate a better overview of an article pair and avoid overlaps across 
text passages and alignments, as such overlaps would contradict the overview goal. 
Note that the interlingual alignment also shapes the text passages, 
i.e. the borders of the aligned text passages are mutually dependent and 
determined during the alignment process to increase the interlingual similarity.
In practice, it is possible that an article contains multiple alternatives 
for an alignment, from which we select only the best matching one in these
settings.
For example,
the information from the first English sentence in Fig. \ref{fig:introduction_example} 
(the construction of the bridge and the official opening) is spread across 
two different German sentences.
This information could be aligned under a different problem 
setting where one would focus on a particular predefined text passage and 
find all possible matching text passages in the other article. 
Such problem variation would require an adaptation of the alignment 
algorithms and is an interesting extension for future work.

\textit{Consecutiveness in the alignment model:} In our
problem statement we also require the consecutiveness of sentences in a text passage. 
This modeling decision is taken in favor of providing an overview of the
overlapping parts and putting complementary information in context.
The structure of the resulting text passages is shaped by the fragmentation of 
similar information in both languages. 
Our model does not require that each sentence in 
a text passage has a 1:1 correspondence in the alignment, such that 
aligned text passages can contain partially overlapping sentences 
or intermediate sentences with no correspondence. 
Consequently, when building pairs of consecutive text passages, 
sentences on one side can contain complementary 
(or sometimes contradictory information) 
and deliver its language-specific context.
An interesting direction for future research is to develop
interlingual Information Extraction methods that would allow precise
identification of the corresponding and additional 
information nuggets within the aligned text passages.

\textit{Cultural studies and Applications:} 
The MultiWiki text passage alignment method presented in this article 
can facilitate and support cross-lingual case studies.
In the context of cultural studies like \cite{Rogers2013} the
proposed text passage alignment approach can reduce the amount of information 
that needs to be manually analyzed by
researchers and enables researches to focus
on the essentially overlapping article parts. A case study with the
digital humanities researchers utilizing
the system for cultural analytics is an interesting extension
for our future work.
The cross-lingual text passage alignment can also enhance a wide range of
interlingual applications that use Wikipedia as an information source.
Example applications that use interlingual Wikipedia content include 
multilingual summarization
\cite{Baralis:2015} and cross-lingual text classification \cite{Ni:2011}.
Our method can provide a more precise context for such applications
through the targeted alignment of the most relevant text passages.
For another example, in our recent demo paper \cite{gottschalk2016}
we presented a novel graphical user interface to 
analyze temporal evolution of partner articles. This
tool uses the interlingual text passage alignment presented in this
article to facilitate a detailed visual article comparison.
As these examples illustrate, interlingual text passage alignment methods
developed in this article have a great potential to facilitate cultural studies
and spawn a
variety of novel interlingual applications.